\documentclass{article}

\usepackage{makecell}
\usepackage{hyperref}
\usepackage{xurl}
\usepackage{xcolor}
\definecolor{crimson}{RGB}{220,20,60}
\usepackage[table]{xcolor}   
\usepackage{cancel}
\usepackage{tikz}

\usepackage{microtype}
\usepackage{graphicx}
\usepackage{subcaption}
\usepackage{booktabs} 

\usepackage{hyperref}

\usepackage[preprint]{icml2026}

\usepackage{amsmath}
\usepackage{amssymb}
\usepackage{mathtools}
\usepackage{amsthm}

\usepackage[capitalize,noabbrev]{cleveref}

\theoremstyle{plain}

\theoremstyle{definition}

\theoremstyle{remark}

\usepackage[textsize=tiny]{todonotes}

\icmltitlerunning{}

\begin{document}

\twocolumn[
\icmltitle{Edge Cluster Expansion with Radial Rotary Attention for Interatomic Potentials}



  \begin{icmlauthorlist}
    \icmlauthor{Zemin Xu}{sch1,sch2}
    \icmlauthor{Wenbo Xie}{sch1}
    \icmlauthor{P. Hu}{sch1}

    Code: \texttt{\href{https://github.com/xvzemin/tace}{\color{crimson}{https://github.com/xvzemin/tace}}} \\
    Models: \texttt{\href{https://github.com/xvzemin/tace-foundations}{\color{crimson}{https://github.com/xvzemin/tace-foundations}}}
  \end{icmlauthorlist}

\icmlaffiliation{sch1}{School of Physical Science and Technology, ShanghaiTech University, Shanghai, China}
\icmlaffiliation{sch2}{School of Chemistry, Nanjing University, Nanjing, China}

  \icmlcorrespondingauthor{Wenbo Xie}{xiewb1@shanghaitech.edu.cn}

  \icmlkeywords{ICT, MLIP, uMLIP}

  \vskip 0.3in
]



\printAffiliationsAndNotice{}  

\begin{abstract}
  In this paper, we provide a systematic investigation of SO(2) theory to machine learning interatomic potentials (MLIPs) and identify the limitations of conventional SO(2) Linear architectures relative to SO(3) Clebsch-Gordan Tensor Products (CGTP). Building on these insights, we propose \textbf{direct Cartesian construction and recursive Clebsch-Gordan construction of Wigner D-matrices} and introduce two novel interaction building blocks. First, we propose the \textbf{Edge Complex Product Basis based on Generalized Asymmetric Contraction}, a new formulation for many-body expansion that directly constructs higher-order interactions on edges through complex-valued equivariant multiplications. Second, we introduce \textbf{Radial Rotary Complex Attention} (RRA), which enhances extrapolation performance and surpasses existing attention vector formulations. We also introduce several improvements to the Atomic Cluster Expansion module. Building on these advances, we train our models on OMat24, sAlex, and MPTrj, and introduce \textbf{TECE-OAM-RRA-1.0}, which achieve state-of-the-art (SOTA) performance on the Matbench Discovery.
\end{abstract}

\section{Introduction}
  Equivariant machine learning interatomic potentials (MLIPs) have become indispensable tools for molecular dynamics simulations. However, the high computational cost of the Clebsch-Gordan tensor product (CGTP) has motivated the development of alternative tensor product formulations aimed at reducing computational complexity. Representative examples include the grid-based Gaunt Tensor Product (GTP)~\cite{GTP}, which exhibits a path-merging property that avoids the explicit evaluation of multiple tensor product paths. Building upon this formulation, EquiformerV3~\cite{eqv3} incorporates nonlinearity together with many-body interactions~\cite{MACE}, resulting in the SwiGLU-S$_2$ architecture. Despite its computational efficiency, GTP inherently excludes antisymmetric interactions and introduces equivariance errors, although these limitations appear to have limited impact on practical fitting performance. The Matrix Tensor Product~\cite{e3x} similarly exploits path merging while accelerating computation through modified contraction schemes, providing an elegant and efficient simplification of conventional tensor products. By explicitly rotating features into an edge-aligned local frame, the SO(2) Linear framework not only provides a sparse implementation of CGTP operations, but also introduces a fundamentally new class of equivariant operators that go beyond conventional tensor products. This approach exhibits improved scaling behavior and enables additional operations on edges at relatively low computational cost. In contrast, irreducible Cartesian tensor products and contractions~\cite{TACE,c3j} perform equivariant convolutions directly in Cartesian space and have demonstrated favorable efficiency-accuracy trade-offs at low angular degrees. Wigner-6j Convolutions~\cite{e2former} further maximize the fusion of edge operations into node-level computations, resulting in highly memory-efficient convolutions. Finally, Vector Spherical Tensor Product (VSTP) approaches~\cite{vsh,vstp} combine vector spherical harmonics with tensor signals to address the missing antisymmetric interactions in grid-based methods.
  
  Among these emerging tensor product formulations, the SO(2) theory has become the most widely adopted in practice. Nevertheless, the theoretical understanding of SO(2) methods remains incomplete. In particular, it is often observed that under the same angular resolution, SO(2)-based models exhibit lower extrapolation and accuracy than CGTP-based approaches. However, models built upon SO(2) operators demonstrate superior accuracy and faster convergence on the universal MLIP (uMLIP) benchmark. Although SO(2) Linear exhibits more favorable asymptotic scaling behavior in principle, this does not necessarily make it the preferred choice in practice. Beyond the aforementioned limitations above, practical implementations typically require two SO(2) Linear operations together with nonlinear transformations applied at the edge level. This reliance on complex edge-level operations prevents the development of highly optimized, general-purpose edge-fusion libraries analogous to OpenEquivariance~\cite{oeq} and CuEquivariance~\cite{cueq}, which are available for CGTP-based models. In our view, the modest accuracy gains observed in uMLIP do not justify the additional implementation complexity and the loss of fusion opportunities introduced by these edge-level nonlinear operations. Conversely, if such nonlinear edge operations are omitted, the advantages of SO(2) Linear largely disappear, and one can simply employ CGTP directly without introducing the additional complexity of SO(2) operators. \textbf{As we will demonstrate in subsequent experiments, the reduced extrapolation capability primarily arises from model-design choices rather than from the SO(2) operations themselves. In particular, an appropriate treatment of radial information enables a substantially broader model-design space without compromising extrapolation performance. By contrast, the accuracy improvements observed on the uMLIP benchmark can largely be attributed to the use of edge nonlinearities and higher edge body orders.} In this work, we revisit the complex-valued nature of SO(2) equivariance. From this perspective, we analyze the design principles and constraints of learnable weights under complex-valued representations and characterize the class of operations that preserve SO(2) equivariance. In the experimental section, we first introduce a faster and numerically more stable method for constructing Wigner-D matrices by leveraging the relationship between spherical tensors and irreducible Cartesian tensors. Next, we compare the behavior of uuSO(2) Linear and uvSO(2) Linear, identifying the regimes in which each formulation exhibits its respective strengths and limitations. We then investigate the impact of complex-valued weight parameterizations in SO(2) Linear operators by examining three representative schemes: \texttt{w1}, \texttt{w1\_w1}, and \texttt{w1\_w2}. Building upon the SO(2) Tensor Product, i.e., complex-valued equivariant multiplication, we introduce an edge cluster expansion (ECE) block that performs many-body expansions directly on edges. In this part, we further demonstrate the advantages of ECE and SO(2) operations through experiments on M-fold rotationally symmetric structures~\cite{mfold} and k-body counterexamples~\cite{Incompleteness}. In addition, we show how complex-conjugate multiplication can be used to construct an SO(2)-invariant ($m=0$, complex-valued) from query and key representations, which naturally leads to the formulation of Radial Rotary Attention. Finally, we evaluate the proposed designs through experiments on the Matbench Discovery benchmark~\cite{Matbench}.

\section{Preliminaries}

\subsection{SO(3) Irreps $\leftrightarrow$ SO(2) Irreps} 
  The theoretical framework related to SO(3)/O(3) operations in machine learning was first systematically implemented in e3nn~\cite{e3nn1,e3nn2,e3nn3}. Concretely, an irreducible representation (irrep) of SO(3) is indexed by the degree $\ell$ and order $m$, and learnable features can be represented as a tensor of dimension $2\ell+1$. In machine learning applications, an additional channel dimension $c$ is typically introduced. Therefore, a complete SO(3) irrep is indexed by $(c, \ell, m)$. In general, different values of $\ell$ may correspond to different numbers of channels; however, in this work, we do not explicitly distinguish between them. 
  The concept of SO(2)-related representations was first introduced in eSCN~\cite{eSCN}, where the authors defined the SO(2) Linear operation. Subsequently, QHNetV2~\cite{QHNetV2} further developed the theoretical framework of SO(2) for Hamiltonian prediction tasks, introducing operations such as SO(2) Tensor Product and SO(2) Gate. The SO(2) group, which describes rotations in the plane, is abelian. Therefore, over the complex numbers, all its irreducible representations are $1$-dimensional. For SO(2) features, the action of a rotation by $\theta$ is given by multiplying features by the unit complex number $e^{i m \theta}, m \in \mathbb{Z}$,
  \begin{equation}
    \mathbf{x_m} \mapsto e^{i m \theta} \mathbf{x_m}.
  \end{equation}
  In geometric graph neural networks, for edge-level SO(3) features, if we rotate all edges to a common reference direction and apply the corresponding Wigner-D transformation to the features, the rotational degrees of freedom reduce to a single remaining angle. In this case, the features can be regarded as local SO(2) features, allowing us to apply SO(2)-related operations (from global SO(3) frame to local SO(2) frame). Similarly, by applying the transpose of the Wigner-D matrix, the features can be transformed back. 

\subsection{SO(2) Operations}
  In general, complex-valued SO(2) features are obtained by rotating real-valued SO(3) features. However, operations on SO(2) features are more naturally interpreted from the perspective of complex numbers. In this work, all SO(3) features are treated in the real-valued representation, while all SO(2) features are interpreted in the complex-valued representation. In practice, we can use \texttt{torch.view\_as\_complex} and \texttt{torch.view\_as\_real}~\cite{torch} to convert between real-valued and complex-valued features. 
  
  \textbf{Given complex-valued weights $\mathbf{w} = w_1 + i w_2$ and complex-valued features $\mathbf{x_m} = x_{+m} + i x_{-m}$, complex multiplication $\mathbf{w} \cdot \mathbf{x_m}$ commutes with rotations and is therefore SO(2)-equivariant. Specifically, for a complex-linear map $L_W(\mathbf{x_m})=W\mathbf{x_m}$, we have
  \begin{equation}
  \label{eq:linear_map}
    L_W\!\left(e^{\mathrm{i}m\theta}\mathbf{x_m}\right)
    =
    We^{\mathrm{i}m\theta}\mathbf{x_m}
    =
    e^{\mathrm{i}m\theta}W\mathbf{x_m}
    =
    e^{\mathrm{i}m\theta}L_W(\mathbf{x_m}).
  \end{equation}
  This implies that we simultaneously modulate both the magnitude and the phase of the complex-valued features. In particular, if we impose the constraint $w_1 = w_2$, this implies that we only apply a fixed phase rotation together with an arbitrary magnitude transformation. If we instead constrain $w_2 = 0$, then the operation only modulates the magnitude while preserving the phase unchanged. We refer to these three weighting schemes for complex-valued features as \texttt{w1\_w2}, \texttt{w1\_w1}, and \texttt{w1}, respectively.} 

\paragraph{SO(2) Linear.}
  Once the embedding mechanism of complex-valued weights is understood, it becomes immediately clear that the essence of the SO(2) Linear operation is to mix features with the same order $m$ across different channels $c$ and degrees $\ell$ in the complex domain. This operation allows us to \textbf{scalarize information} with higher angular resolution, $(\ell > 0,\, m = 0) \rightarrow (\ell = 0,\, m = 0)$, thereby achieving an effect analogous to an SO(3) tensor product. See Section~\ref{sec:mfoldandkbody} for details.

\paragraph{SO(2) Tensor Product}
  Given two complex-valued SO(2) features $\mathbf{x}_{m_1}$ and $\mathbf{x}_{m_2}$, their complex multiplication remains equivariant under rotations when $m_3 = m_1 + m_2$ or $m_3 = m_1 - m_2$, because
  \begin{equation}
    \mathbf{x}_{m_1} \mathbf{x}_{m_2}
    \mapsto
    e^{i(m_1+m_2)\theta}
    \mathbf{x}_{m_1} \mathbf{x}_{m_2},
  \end{equation}
  and
  \begin{equation}
    \mathbf{x}_{m_1} \overline{\mathbf{x}}_{m_2}
    \mapsto
    e^{i(m_1-m_2)\theta}
    \mathbf{x}_{m_1} \overline{\mathbf{x}}_{m_2},
  \end{equation}
  where $\overline{\mathbf{x}}_{m_2}$ denotes the complex conjugate of $\mathbf{x}_{m_2}$. Although some previous work calls SO(2) Linear a ``tensor product'', it is actually different from the SO(2) Tensor Product. They handle different parts of the model, and both are important.

\paragraph{SO(2) Gate.}
  Different from the SO(3) gate, where the $\ell=0, m=0$ features are used as gates, the SO(2) Gate uses the $m=0$ features from all degrees $\ell$ as gates. This can be regarded as a special case of the SO(2) tensor product, so we do not discuss it in detail.

\subsection{Special Cases of Node-wise SO(2) Operations}
  SO(2) operations are typically performed on edges, where the edge direction naturally provides a local frame. If equivalent operations could be carried out directly on node representations, additional computational savings might be achieved. One possible direction is represented by frame-averaging approaches~\cite{fram_averaging,QHNetV2}, which construct a local reference axis from the target atom and its nearby neighboring source atoms. While such formulations have proven effective for Hamiltonian prediction tasks, their applicability to interatomic potentials is less clear. The selection of a particular neighbor may introduce discontinuities when the local atomic environment changes. Averaging over all neighboring atoms can alleviate this issue; however, the resulting formulation becomes conceptually similar to aggregating edge-wise operations at the node level, thereby diminishing the computational benefits originally sought. A second direction originates from attempts to simplify Wigner-6j Convolutions~\cite{e2formerv2}. Wigner-6j Convolutions exploit solid harmonics and binomial expansions to separate the angular dependence of source and target atoms. Since CGTP remains computationally expensive, the factorized representation enabled by solid harmonics naturally motivates the construction of local reference axes directly from atomic positions. Therefore, although node-wise SO(2) operations constitute a promising design space, a universally applicable design paradigm that generalizes across diverse architectures is still lacking.

\section{Results}

\subsection{O(2) Irrep Notation}
  Although our model is only required to be local SO(2)-equivariant, it is useful to first introduce the notation for irreducible representations of O(2). This notation will be used in the subsequent discussion of Weight Types~\ref{weight_type}, Edge Cluster Expansion~\ref{ece}, and Radial Rotary Attention~\ref{RRA}. Under a planar reflection, complex feature is mapped to its conjugate:
  \begin{equation}
  \label{eq:reflection}
    \mathbf{x_m} \mapsto \overline{\mathbf{x_m}}.
  \end{equation}
  Therefore, for $m>0$, the corresponding $\mathrm{O}(2)$ irrep is still labeled by the order $m$, while the special case is $m=0$. Since the rotation action is trivial, reflection can act either as $+1$ or $-1$, giving two different one-dimensional irreps: $0e$ and $0o$. Thus, the irreps of $\mathrm{O}(2)$ can be summarized as
  \begin{equation}
      0e,\quad 0o,\quad 1m,\quad 2m,\quad 3m,\quad \ldots
  \end{equation}
  where $m>0$ labels two-dimensional irreps, and only the $m=0$ case splits into even and odd parity.

\subsection{Cartesian Construction of Wigner-D Matrices}
  Spherical tensors and Cartesian tensors provide two equivalent descriptions of
  three-dimensional equivariant features. A spherical tensor of degree $\ell$,
  denoted by $\mathbf{x}^{(\ell)} \in \mathbb{R}^{2\ell+1}$, transforms under a
  rotation $\mathbf{R}\in\mathrm{SO}(3)$ according to the Wigner-D matrix:
  \begin{equation}
      \mathbf{x}^{(\ell)\prime}
      =
      \mathbf{D}^{(\ell)}(\mathbf{R})\mathbf{x}^{(\ell)}.
      \label{eq:spherical_rotation}
  \end{equation}
  In contrast, a rank-$\ell$ Cartesian tensor
  $\mathbf{T}^{(\ell)}\in\mathbb{R}^{3\times\cdots\times3}$ transforms by applying
  the rotation matrix independently to each Cartesian index:
  \begin{equation}
      T^{(\ell)\prime}_{i_1\cdots i_\ell}
      =
      \sum_{j_1,\ldots,j_\ell}
      R_{i_1j_1}\cdots R_{i_\ell j_\ell}
      T^{(\ell)}_{j_1\cdots j_\ell}.
      \label{eq:cartesian_rotation}
  \end{equation}
  Here, we propose two methods for computing the Wigner-D matrix from rotation matrices. The first method is more intuitive and easier to understand, while the second provides a more efficient implementation.

\paragraph{Direct Cartesian Construction.}
  The irreducible Cartesian tensor decomposition (ICTD)~\cite{ICTdecomposition} provides an orthonormal basis transformation between degree-$\ell$ spherical tensors and rank-$\ell$ symmetric traceless Cartesian tensors. Let $\mathbf{C}_{\ell} \in \mathbb{R}^{3^\ell\times(2\ell+1)}$ denote the path matrix. The spherical-to-Cartesian and Cartesian-to-spherical transformations are respectively given by
  \begin{equation}
      \operatorname{vec}\!\left(\mathbf{T}^{(\ell)}\right)
      =
      \mathbf{C}_{\ell}\mathbf{x}^{(\ell)},
      \qquad
      \mathbf{x}^{(\ell)}
      =
      \mathbf{C}_{\ell}^{\mathsf T}
      \operatorname{vec}\!\left(\mathbf{T}^{(\ell)}\right).
      \label{eq:ictd_transform}
  \end{equation}
  Applying the Cartesian rotation in Eq.~\eqref{eq:cartesian_rotation} between the two basis transformations gives a direct construction of the Wigner-D matrix:
  \begin{equation}
      \mathbf{D}^{(\ell)}(\mathbf{R})
      =
      \mathbf{C}_{\ell}^{\mathsf T}
      \mathbf{R}^{\otimes\ell}
      \mathbf{C}_{\ell}.
      \label{eq:cartesian_wigner}
  \end{equation}
  \paragraph{Limitations of the Direct Cartesian Construction.}
    Although Eq.~\eqref{eq:cartesian_wigner} is numerically stable, its direct evaluation introduces a Cartesian intermediate tensor containing $3^\ell$ components. When evaluated for a batch of $E$ edges, the intermediate storage scales as $\mathcal{O}\left(E\,3^\ell(2\ell+1)\right)$. Consequently, the direct construction becomes increasingly expensive as $\ell$ grows, despite the final Wigner-D block containing only $(2\ell+1)^2$ entries.

\paragraph{Recursive Clebsch-Gordan Construction.}
  \begin{figure}[h]
    \begin{center}
      \centerline{\includegraphics[width=\columnwidth]{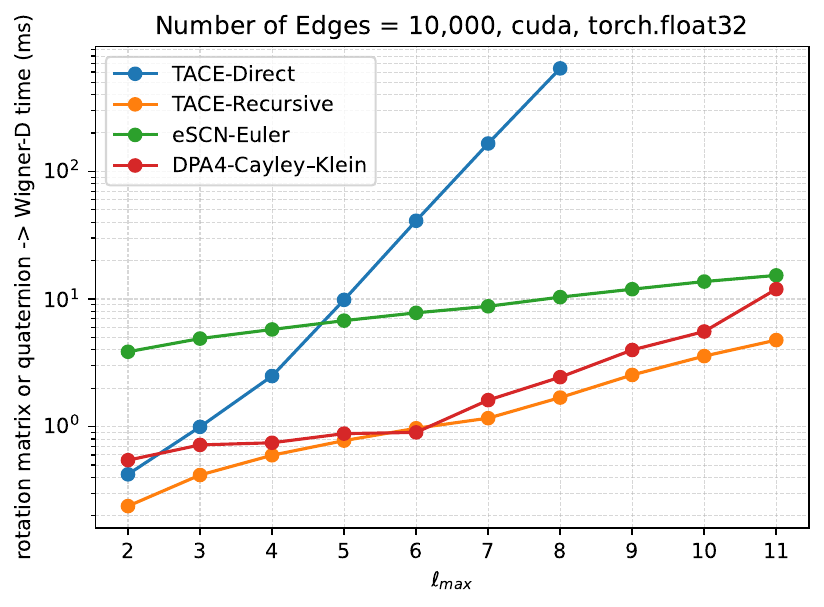}}
      \caption{
        Runtime comparison of different Wigner-D matrix construction methods from rotation matrices or quaternion. The benchmark is performed on an NVIDIA RTX 4090 GPU using 10000 edges,
        torch.float32 precision, and averages the runtime over repeated forward evaluations.
        The recursive Clebsch-Gordan construction shows better scaling with increasing $\ell_{\max}$ (construct from 0 to $\ell_{\max}$)
        compared with other methods.
      }
      \label{fig:wignerD-benchmark}
    \end{center}
  \end{figure}
  We note that the path matrix itself is generated recursively by coupling a degree-one representation with a degree-$(\ell-1)$ representation $1\otimes(\ell-1)\rightarrow\ell$. This structure allows us to avoid the explicit rank-$\ell$ Cartesian tensor. Let
  \begin{equation}
      \mathbf{K}^{(\ell)}_{abM}
      =
      \operatorname{Wigner3j}(1,\ell-1,\ell)_{abM}
      \label{eq:recursive_cg}
  \end{equation}
  denote the fixed Clebsch-Gordan coupling coefficients, where
  $a\in\{1,2,3\}$, $b\in\{1,\ldots,2\ell-1\}$, and
  $M\in\{1,\ldots,2\ell+1\}$. Starting from
  \begin{equation}
      \mathbf{D}^{(0)}(\mathbf{R}) = 1,
      \qquad
      \mathbf{D}^{(1)}(\mathbf{R}) = \mathbf{R},
      \label{eq:recursive_base}
  \end{equation}
  higher-degree Wigner-D matrices can be constructed recursively as
  \begin{equation}
      D^{(\ell)}_{MN}(\mathbf{R})
      =
      (2\ell+1)
      \sum_{a,b,c,d}
      K^{(\ell)}_{abM}
      D^{(1)}_{ac}(\mathbf{R})
      D^{(\ell-1)}_{bd}(\mathbf{R})
      K^{(\ell)}_{cdN}.
      \label{eq:recursive_wigner}
  \end{equation}
  The factor $2\ell+1$ accounts for the normalization convention of the Wigner-$3j$ symbols. Equation~\eqref{eq:recursive_wigner} is algebraically equivalent to Eq.~\eqref{eq:cartesian_wigner}, but it avoids the exponentially large Cartesian representation. The largest intermediate tensor in the recursive contraction scales as $\mathcal{O} \left(E\,(2\ell-1)(2\ell+1)\right)$ which replaces the exponential scaling with polynomial scaling. Moreover, the recursive method has lower numerical error and avoids the inverse trigonometric functions and angular singularities present in the original eSCN formulation. It consists only of fixed-shape tensor contractions, making it suitable for specialized GPU kernels. For the asymptotic behavior of different methods, please refer to Fig~\ref{fig:wignerD-benchmark}. As shown in the figure, our method achieves the best asymptotic performance. Notably, UMA~\cite{UMA} and DPA4~\cite{DPA4} constructs the Wigner-D matrices using quaternions and custom operators implemented in more than 1000 lines of code. In contrast, our method requires fewer than 100 lines of code. Since the accuracy of the initial rotation matrix $R$ is also crucial for the recursive construction, we likewise construct $R$ from quaternions.
  \begin{algorithm}[tb]
    \caption{Recursive Construction of Wigner-D Matrices}
    \label{alg:recursive_wigner}
    \begin{algorithmic}
      \STATE {\bfseries Input:} rotation matrices
      $\mathbf{R}\in\mathbb{R}^{E\times3\times3}$,
      maximum degree $L$
      \STATE {\bfseries Output:} Wigner-D blocks
      $\{\mathbf{D}^{(\ell)}\}_{\ell=0}^{L}$
      \STATE Initialize
      $\mathbf{D}^{(0)} \leftarrow \mathbf{1}$
      \IF{$L \geq 1$}
        \STATE $\mathbf{D}^{(1)} \leftarrow \mathbf{R}$
      \ENDIF
      \FOR{$\ell=2$ {\bfseries to} $L$}
        \STATE Load coefficients $\mathbf{K}^{(\ell)} \leftarrow \operatorname{Wigner3j}(1,\ell-1,\ell)$
        \STATE Compute the next Wigner-D block by one contraction:
        \begin{equation*}
        D^{(\ell)}_{eMN}
        \leftarrow
        (2\ell+1)
        \sum_{a,b,c,d}
        K^{(\ell)}_{abM}
        D^{(1)}_{eac}
        D^{(\ell-1)}_{ebd}
        K^{(\ell)}_{cdN}.
        \end{equation*}
      \ENDFOR
      \STATE Return $\{\mathbf{D}^{(\ell)}\}_{\ell=0}^{L}$
    \end{algorithmic}
  \end{algorithm}

\subsection{SO(3) Paths and SO(2) Paths}
  For SO(3) operations, we typically refer to a tuple $(\ell_{\text{in1}}, \ell_{\text{in2}}, \ell_{\text{out}})$ as a \emph{path}. 
  Each path is associated with a learnable weight parameter. The weights are defined with respect to the output degree $\ell_{\text{out}}$, while the feature mixing over $\ell_{\text{in1}}$ and $\ell_{\text{in2}}$ is performed through a linear layer.

  For SO(2) operations, in this work, we refer to a tuple
  $(\ell_{\mathrm{in}}, \ell_{\mathrm{out}}, m)$ as a \emph{path}. 
  Similarly, each path is associated with a learnable weight.

\subsection{uvSO2Linear and uuSO2Linear}
  To the best of our knowledge, all existing SO(2) architectures currently adopt the SO(2) Linear introduced in eSCN~\cite{eSCN}. However, considering that this operation is intended to replace tensor products with spherical harmonics, we believe that the correspondence between the two is not entirely straightforward. In conventional SO(3) frameworks, radial information is typically associated with the SO(3) paths arising from the uuu SO(3) tensor product, with each path receiving its own path-dependent weighting. In contrast, under the SO(2) framework, radial information is usually incorporated before the SO(2) Linear operation through SO(3) real multiplication or SO(2) complex multiplication where the radial weights are real-valued (i.e., with zero imaginary part). As a result, radial parameters are generally not directly constructed for each individual SO(2) path. We therefore argue that this mechanism more closely resembles uvw SO(3) tensor product, where radial dependencies cannot be incorporated directly for each path due to the large number of paths. Therefore, we observe that the SO(2) Linear layer introduced in eSCN is, in essence, a fully connected linear transformation that does not rely on path-wise radial weights. By analogy with the terminology used in SO(3) tensor product, we refer to this operation as \texttt{uvSO2Linear}. Therefore, we argue that the truly ''equivalent`` counterpart of the uuu SO(3) tensor product should instead be \texttt{uuSO2Linear}. For a fair comparison, we use a single SO(2) Linear operation throughout and do not introduce any edge-wise nonlinearities. Under this formulation, we compare the performance of the two approaches with the same number of parameters. As shown in Table~\ref{tab:uuSO2Linear}, \texttt{uuSO2Linear} achieves accuracy and extrapolation performance comparable to those of uuu SO(3) tensor product. In contrast, \texttt{uvSO2Linear}, which does not directly depend on radial information, exhibits a noticeable loss in expressive power.

  \paragraph{Advantages and Limitations.}
    By more effectively leveraging radial information, \texttt{uuSO2Linear} offers stronger expressive power and extrapolation capabilities. However, this formulation does not permit channel mixing. In contrast, \texttt{uvSO2Linear} allows arbitrary channel transformations. When sufficient training data are available, the improvements in accuracy and extrapolation offered by \texttt{uuSO2Linear} may become marginal. In this regime, the arbitrary channel transformations enabled by \texttt{uvSO2Linear} can be combined with edge nonlinearities, making it more flexible from an architectural perspective.

    We consider \texttt{uuSO2Linear} important because, when designing operators in an SO(2) framework, a single linear layer should ideally serve as a replacement for the \texttt{uuuSO3TensorProduct}. While both SO(2) Linear formulations are theoretically valid, using \texttt{uvSO2Linear} in this setting sacrifices compatibility with edge nonlinearities. On the other hand, introducing such nonlinearities precludes the memory-efficient optimization of fusing edge computations.

\begin{table}[tb]
\caption{Comparison of the extrapolation performance of different SO(2) Linear type on the 3BPA and AcAc datasets, using CGTP as the baseline. Root mean square errors (RMSEs) are reported for energies (E) in meV and forces (F) in meV/\AA. The best results are highlighted in bold.}
  \label{tab:uuSO2Linear}
  \begin{center}
    \begin{small}
        \begin{tabular}{l|ccc|ccc}
          \toprule
          Model & \multicolumn{6}{c}{TACE (only one SO(2) Linear layer)}\\
          \toprule
          Dataset  & \multicolumn{3}{c|}{3BPA} & \multicolumn{3}{c}{AcAc} \\
          \midrule
          Type &  uu  & uv & cgtp  & uu & uv & cgtp\\
          \midrule
          300K\phantom{0000}E ($\downarrow$) & \textbf{3.7} & 4.7  & 3.8  & \textbf{1.0}  & 1.5 & 1.2 \\
          \phantom{300K0000}F ($\downarrow$) & \textbf{8.8}  & 11.8 & 9.1  & \textbf{5.2}  & 6.8 & 5.4 \\
          600K\phantom{0000}E ($\downarrow$) & \textbf{9.1}  & 15.2 & 10.5 & \textbf{5.2}  & 5.4 & 5.6 \\
          \phantom{600K0000}F ($\downarrow$) & \textbf{20.5} & 27.5 & 21.5 & \textbf{22.8} & 28.3& 23.5\\
          1200K\phantom{000}E ($\downarrow$) & \textbf{27.9} & 30.7 & 29.5 & - & - & - \\
          \phantom{1200K000}F ($\downarrow$) & \textbf{59.0} & 70.4 & 62.0 & - & - & - \\
          dihedral\phantom{00}E ($\downarrow$) & 15.0 & \textbf{14.5} & 15.6 & - & - & - \\
          \phantom{dihedral00}F ($\downarrow$) & \textbf{18.1} & 22.2 & 20.2 & - & - & - \\
          Params\phantom{000}(M) & 5.7 & 5.7 & 5.7 & 4.9 & 4.9 & 4.9 \\
          \bottomrule
        \end{tabular}
    \end{small}
  \end{center}
\end{table}

\begin{table*}[tb]
\caption{Comparison of the extrapolation performance of different SO(2) Linear weight types and attention mechanisms on the 3BPA dataset. Root mean square errors (RMSEs) are reported for energies (E) in meV and forces (F) in meV/\AA. Within each attention mechanism, the best-performing architecture is highlighted in bold. Among the three attention mechanisms, the overall best-performing one is highlighted in {\setlength{\fboxsep}{0pt}\colorbox{green!10}{\strut green}}.}
  \label{tab:RRCA}
  \begin{center}
    \begin{small}
        \begin{tabular}{l|ccc|ccc|ccc}
          \toprule
          Model & \multicolumn{9}{c}{TACE (uvSO2Linear + SO(2)Gate + uvSO2Linear + Attention), 2 Layer, $\nu=2$}\\
          \toprule
          Attention Type & \multicolumn{3}{c|}{Radial Rotary Complex Attention} & \multicolumn{3}{c|}{EquiformerV3 Attention Vector} & \multicolumn{3}{c}{No Attention}\\
          \midrule
          Weight Type & w1\phantom{\_w1}  & w1\_w1 & w1\_w2  & w1\phantom{\_w1}  & w1\_w1 & w1\_w2& w1\phantom{\_w1} & w1\_w1 & w1\_w2\\
          \midrule
          300K\phantom{0000}E ($\downarrow$) &\cellcolor{green!10}\textbf{3.6\phantom{0}}  & 4.2  & 7.1  & 4.6   & \textbf{4.0} &  6.3    & \textbf{4.4} & 4.5 & \textbf{4.4} \\
          \phantom{300K0000}F ($\downarrow$) & \cellcolor{green!10}\textbf{8.3\phantom{0}}  & 10.3 & 12.0 & \textbf{8.9}   & 10.4& 12.7   & \textbf{8.7} & 11.5 & 12.6  \\
          600K\phantom{0000}E ($\downarrow$) & \cellcolor{green!10}\textbf{8.6\phantom{0}}  & 10.7 & 12.9 & \textbf{9.9}   & 10.0& 13.4   & \textbf{9.6} & 12.3& 13.5 \\
          \phantom{600K0000}F ($\downarrow$) & \cellcolor{green!10}\textbf{19.8} & 24.3 & 28.1 & \textbf{20.9} & 27.0 & 30.8 & \textbf{22.3}& 27.0& 29.7 \\
          1200K\phantom{000}E ($\downarrow$) & \cellcolor{green!10}\textbf{27.0} & 34.9 & 41.5 & \textbf{30.3} & 38.5 & 48.6 & 46.3& 41.0&  \textbf{40.5}\\
          \phantom{1200K000}F ($\downarrow$) & \cellcolor{green!10}\textbf{55.4} & 65.9 & 73.5 & \textbf{78.2} & 85.4 & 100.6& 88.8 & \textbf{78.7} & 80.5 \\
          dihedral\phantom{00}E ($\downarrow$) & \cellcolor{green!10}\textbf{10.2} & 10.7 & 21.0 & \textbf{15.0} & 25.0 & 32.2 & 19.4& 23.6& \textbf{18.4} \\
          \phantom{dihedral00}F ($\downarrow$) & \cellcolor{green!10}\textbf{15.1} & 18.4 & 23.7 & \textbf{19.5} & 24.2 & 29.4 & \textbf{19.7} &24.0 & 23.9 \\
          \bottomrule
        \end{tabular}
    \end{small}
  \end{center}
\end{table*}

\subsection{Complex-Valued Weights Break O(2)/O(3) Equivariance and Weaken Extrapolation}
  \label{weight_type}
  We seek to understand why complex-valued weights arise in force-field models whose inputs and outputs are both real-valued. This naturally raises the question of whether the complex phase provides meaningful modeling capacity or merely introduces redundant degrees of freedom. To investigate this issue, we compare the accuracy and extrapolation performance of three different SO(2) weight types. The comparison is conducted under three architectures: Radial Rotary Attention, the attention-vector formulation adopted by EquiformerV3~\cite{eqv3}, and a non-attention architecture. For each architecture, the best result is highlighted in \textbf{bold} in Table~\ref{tab:RRCA}. From the results, we observe that introducing phase variations provides little benefit on this dataset. In several cases, it even leads to degradation in both accuracy and extrapolation performance.

  In general, complex-valued weights do not guarantee $\mathrm{O}(2)$-equivariance.
  Taking the \texttt{w1\_w2} as an example, its complex weight is written as
  $w=w_1+\mathrm{i}w_2$. Under the reflection $-I$, the complex SO(2) feature
  $\mathbf{x}_m$ is conjugated, namely $(-I)\mathbf{x}_m=\overline{\mathbf{x}_m}$.
  Therefore, 
  \begin{equation}
  \begin{aligned}
      L_w\!\left((-I)\mathbf{x}_m\right)
      &=
      (w_1+\mathrm{i}w_2)\overline{\mathbf{x}_m}, \\
      (-I)L_w(\mathbf{x}_m)
      &=
      (w_1-\mathrm{i}w_2)\overline{\mathbf{x}_m}.
  \end{aligned}
  \end{equation}
  These two expressions are not equal unless $w_2=0$. Thus, \texttt{w1\_w2} is generally only
  SO(2)-equivariant, while the real-weight \texttt{w1} case is compatible with O(2) reflection symmetry. This implies that, restricting the model to use only real-valued weights allows us to enforce the correct behavior by construction. In contrast, an SO(3)-equivariant model does not impose this constraint and can only rely on the training data to learn the appropriate behavior under global inversion. With sufficiently large and diverse datasets, the model may approximately learn this property from data. However, in the low-data regime, the lack of an explicit constraint becomes much more pronounced, leading to a clear performance gap.

\subsection{Generalized Symmetric Contraction and Uncoupled Edge Cluster Expansion}
\label{ece}

  The Transformer architecture has achieved tremendous success in large language models (LLMs). However, we remain cautious about its direct application to machine learning interatomic potentials. There are two main reasons for this. First, adopting a Transformer architecture typically requires complicated operations explicitly at the edge level. For conventional CGTP frameworks, this can become inefficient and may prevent the direct use of mature operator-fusion libraries such as OpenEquivariance~\cite{oeq} and CuEquivariance~\cite{cueq}. Second, the success of models such as NequIP~\cite{NequIP}, MACE~\cite{MACE}, EquFlash~\cite{equflashv1}, and TACE~\cite{TACE} shows that strong performance can already be achieved without graph softmax mechanisms.

  In light of expressive Transformer-based models such as EquiformerV3~\cite{eqv3} and PET~\cite{PET}, we believe that explicitly modeling source-target atom relationships during large-scale pretraining is at least unlikely to be harmful. Therefore, we introduce edge-level many-body interactions through an uncoupled edge cluster expansion (ECE). The key idea is to perform tensor contractions directly on edge features using generalized coupling coefficients. The symmetric contraction was first introduced in MACE~\cite{MACE} with O(3) equivariance. Here, we generalize this construction and extend it to SO(2). More broadly, the framework is not limited to SO(2) and O(3): any tensor representation equipped with Clebsch-Gordan-like coupling coefficients can be incorporated into the same formulation. We write the input features in layout as $\mathbf{x}^{(r)} \in \mathbb{R}^{B \times F \times C}$, where $B$ is the batch size, $F$ is the feature dimension of the representation, and $C$ is the channel dimension. For SO(2) features, $l$ index are merged into channel dimension. 

  \paragraph{Generalized Coupling Coefficients.}
  Consider a representation decomposed into blocks indexed by $\lambda$, where each block has dimension $d_\lambda$. For O(3), $\lambda$ may correspond to $(\ell,p)$; for SO(2), $\lambda=m$. A pairwise coupling is described by coefficients
  \begin{equation}
      C^{\lambda_{\mathrm{out}}}_{\lambda_1\lambda_2}
      \in
      \mathbb{R}^{d_{\lambda_1}\times d_{\lambda_2}\times d_{\lambda_{\mathrm{out}}}}.
  \end{equation}
  More generally, a correlation order $\nu$ coupling path is denoted as
  \begin{equation}
      \eta_\nu
      =
      (\lambda_1,\lambda_2,\ldots,\lambda_\nu;
      \lambda_{\mathrm{out}}),
  \end{equation}
  with generalized coefficients
  \begin{equation}
      \mathcal{C}^{\lambda_{\mathrm{out}}}_{\eta_\nu}
      \in
      \mathbb{R}^{d_{\lambda_1}\times\cdots\times d_{\lambda_\nu}
      \times d_{\lambda_{\mathrm{out}}}}.
  \end{equation}
  These coefficients can be constructed recursively from pairwise couplings, as in the generalized Clebsch-Gordan construction used by MACE~\cite{MACE}. This makes symmetric contraction a general algebraic operation rather than a construction specific to O(3).

  \paragraph{SO(2) as a Special Case.}
  For SO(2), the representation in real basis has
  \begin{equation}
      d_m =
      \begin{cases}
          1, & m=0,\\
          2, & m>0.
      \end{cases}
  \end{equation}
  The allowed pairwise output frequencies follow from complex multiplication:
  \begin{equation}
      m_{\mathrm{out}} = m_1 + m_2,
      \qquad
      m_{\mathrm{out}} = |m_1-m_2|.
  \end{equation}
  For example, for the sum path with $m_1,m_2>0$,
  \begin{equation}
  \begin{aligned}
      (a+\mathrm{i}b)(c+\mathrm{i}d)
      =
      (ac-bd)+\mathrm{i}(ad+bc),
  \end{aligned}
  \end{equation}
  which gives the nonzero real basis coefficients
  \begin{equation}
      C_{000}=1,\quad
      C_{110}=-1,\quad
      C_{011}=1,\quad
      C_{101}=1.
  \end{equation}
  To maintain stable output variance, we normalize the coefficients:
  \begin{equation}
      \sum_{a,b}
      \left(
      C^{m_{\mathrm{out}}}_{m_1m_2,abo}
      \right)^2
      =
      1.
  \end{equation}

  \paragraph{Limitations of Symmetric Contraction.}
  The standard symmetric contraction repeatedly contracts the same feature:
  \begin{equation}
      \mathbf{x}
      \otimes
      \mathbf{x}
      \otimes
      \cdots
      \otimes
      \mathbf{x}.
  \end{equation}
  This has two limitations.   First, symmetric contraction is not always variance stable. Even if the Clebsch-Gordan coefficients are normalized, repeatedly using the same feature tensor introduces correlations between different polynomial orders, such as $\mathbf{x}$, $\mathbf{x}^2$, and $\mathbf{x}^3$. As a result, the variance of the summed output cannot be normalized in a straightforward manner. This effect becomes more severe for high-order expansions, large numbers of paths, or extreme local structures, and may lead to unstable training. Second, symmetric features inherently admit fewer computational paths than asymmetric implementations. In addition, the symmetry of the Clebsch-Gordan coefficients causes antisymmetric components to vanish automatically under O(3). Although the contributions from higher-order terms gradually decrease, and symmetric contraction may appear more intuitive, our objective is to perform a many-body expansion. Therefore, it is necessary to relax these constraints appropriately.

  \paragraph{Generalized Asymmetric Contraction.}
  To relax these constraints, we introduce asymmetric contraction. The input can be a list of feature tensors
  \begin{equation}
      \left[
      \mathbf{x}^{(1)},
      \mathbf{x}^{(2)},
      \ldots,
      \mathbf{x}^{(\nu_{\max})}
      \right].
  \end{equation}
  If only one tensor is provided, it is reused for all orders. If fewer than $\nu_{\max}$ tensors are provided, the last tensor is reused for the remaining orders. For a path $\eta_\nu$, the product basis is
  \begin{equation}
  \mathbf{B}^{(\nu)}_{b,\eta_\nu,c,o}
  =
  \sum_{a_1,\ldots,a_\nu}
  \mathcal{C}^{\lambda_{\mathrm{out}}}_{\eta_\nu,a_1\cdots a_\nu o}
  \prod_{r=1}^{\nu}
  \mathbf{x}^{(r)}_{b,\lambda_r,c,a_r}.
  \end{equation}
  The coefficients may be shared internal weights $\mathbf{W}^{(\nu)} \in \mathbb{R}^{N_{\mathrm{path},\nu}\times C}$ or externally provided edge/node-dependent (non)linear weights $\mathbf{W}^{(\nu)} \in \mathbb{R}^{B\times N_{\mathrm{path},\nu}\times C}$.
  The output is accumulated as
  \begin{equation}
  \mathbf{y}_{b,\lambda,c,o}
  =
  \sum_{\nu=1}^{\nu_{\max}}
  \sum_{\eta_\nu\rightarrow \lambda}
  \frac{1}{
  \sqrt{
  N_{\mathrm{p}}(\lambda,\nu)
  N_{\mathrm{o}}(\lambda)
  }}
  \mathbf{W}^{(\nu)}_{b,\eta_\nu,c}
  \mathbf{B}^{(\nu)}_{b,\eta_\nu,c,o},
  \end{equation}
  where $N_{\mathrm{p}}(\lambda,\nu)$ is the number of paths for order $\nu$ contributing to output block $\lambda$, and $N_{\mathrm{o}}(\lambda)$ is the number of correlation orders contributing to $\lambda$. This keeps the output variance close to one when different input branches are approximately independent. We refer to the recursive algorithm introduced in MACE~\cite{MACE} as the Horner-recursive implementation. It constructs features of successive correlation orders recursively by contracting the input features with a batched version of the generalized coupling coefficients associated with the highest correlation order. However, the size of the batched coefficient scales as $F^{\nu}$. Consequently, the intermediate tensors also scale as $F^{\nu}$, which becomes prohibitively expensive when $F$ is large. For example, in the Cartesian representation, the feature dimension at angular degree $\ell$ grows as $3^\ell$, resulting in a total dimension of $F_{\mathrm{cart}}=\sum_{\ell=0}^{\ell_{\max}}3^\ell$. This exponential growth quickly makes the computation impractical and largely defeats the purpose of the recursive formulation. In contrast, the spherical representation contains $2\ell+1$ components at each angular degree $\ell$, giving a total dimension of $F_{\mathrm{sph}}=\sum_{\ell=0}^{\ell_{\max}}(2\ell+1)=(\ell_{\max}+1)^2$. Under the SO(2) basis, only order with $m\leq m_{\max} \text{or } \ell_{\max}$ are retained, and the corresponding dimension is $F_{\mathrm{SO(2)}}=1+2m_{\max}$. Therefore, this approach is most suitable for node-level operations, or for settings in which memory consumption is not a limiting factor, where it can provide a substantial computational speedup. This comparison also highlights the potential advantage of SO(2) ACE. However, as discussed below, there is currently no general node-level construction for SO(2) because of the discontinuity issue. Consequently, the Horner-recursive implementation, whose intermediate tensors scale exponentially with the correlation order, cannot be directly applied to edge-level operations. We therefore design a memory-efficient algorithm, presented in Algorithm~\ref{alg:generalized_asymmetric_contraction}, which avoids materializing the full high-order intermediate tensors. This design trades computational speed for substantially reduced memory consumption. It is particularly well suited to ECE, where the feature dimension remains small.
  \begin{algorithm}[tb]
    \caption{Generalized Asymmetric Contraction}
    \label{alg:generalized_asymmetric_contraction}
    \begin{algorithmic}
      \STATE {\bfseries Input:} features
      $\{\mathbf{x}^{(r)}\}_{r=1}^{\nu_{\max}}$,
      $\mathbf{x}^{(r)}\in\mathbb{R}^{B\times F\times C}$,
      generalized coefficients $\{\mathcal{C}^{(\nu)}\}$,
      weights $\{\mathbf{W}^{(\nu)}\}$
      \STATE {\bfseries Output:} contracted features
      $\mathbf{y}\in\mathbb{R}^{B\times F_{\mathrm{out}}\times C}$

      \STATE Decompose the feature dimension into representation blocks
      $\{\lambda\}$
      \STATE Initialize $\mathbf{y}\leftarrow 0$

      \FOR{$\nu=1$ {\bfseries to} $\nu_{\max}$}
        \FOR{each ordered path
        $\eta_\nu=(\lambda_1,\ldots,\lambda_\nu;\lambda_{\mathrm{out}})$}
          \STATE Compute $\mathbf{B}^{(\nu)}_{\eta_\nu}$ using
          $\mathcal{C}^{\lambda_{\mathrm{out}}}_{\eta_\nu}$
          \STATE Accumulate
          \[
          \mathbf{y}_{\lambda_{\mathrm{out}}}
          \leftarrow
          \mathbf{y}_{\lambda_{\mathrm{out}}}
          +
          s_{\lambda_{\mathrm{out}},\nu}
          \mathbf{W}^{(\nu)}_{\eta_\nu}
          \mathbf{B}^{(\nu)}_{\eta_\nu}
          \]
        \ENDFOR
      \ENDFOR

      \STATE Merge all output blocks into
      $\mathbb{R}^{B\times F_{\mathrm{out}}\times C}$
      \STATE Return $\mathbf{y}$
    \end{algorithmic}
  \end{algorithm}
  The use of ECE allows the edge basis to depend either on an individual source or target atom, as discussed in Section~\ref{sec:mfoldandkbody}, or jointly on both the source and target atoms. Here, we focus on the latter formulation. Since many-body expansions are typically implemented using channel-wise operations, all angular orders $m$ are required to share the same number of channels. This requirement can be satisfied either by applying uvSO2Linear or padding. More specifically, given the edge-aligned source and target features,
  \(\mathbf{x}_{j}^{t-1}\) and \(\mathbf{x}_{i}^{t-1}\), from the \((t-1)\)-th layer,
  we construct the first-order edge basis, \({}^{\nu=1}\mathbf{E}_{ij}^{t}\),
  at the \(t\)-th layer as
  \begin{equation}
  {}^{\nu=1}\mathbf{E}_{ij}^{t}
  =
  \text{uvSO2Linear}
  \left(
  \mathbf{x}_{i}^{t-1}
  \;\Vert\;
  \mathbf{x}_{j}^{t-1}
  \right),
  \end{equation}
  where $\Vert$ denotes channel-wise concatenation.
  Besides introducing complex-valued weights, we also have what appears to be an additional option: allowing the real and imaginary parts of the weight to depend separately on the source and target atoms. However, as discussed above, we believe that real-valued weights are preferable. Therefore, the greatest flexibility of ECE lies in the choice of edge-dependent real weights.
  \begin{equation}
  \tikz[baseline=(eq.base)]{
    \node[inner sep=0pt] (eq) {$
      \underbrace{w_1}_{\text{source atoms}}
      + \quad
      i\underbrace{w_2}_{\text{target atoms}}
    $};
    \draw[red, line width=0.8pt] 
      ([yshift=5pt]eq.west) -- ([yshift=5pt]eq.east);
  }
  \end{equation}

\paragraph{Nonlinearity and Extrapolation.}
\label{par:extrapolation}
  As one of the early influential equivariant models, NequIP~\cite{NequIP} applies a gating operation after message aggregation. Subsequently, BoTNet~\cite{BotNet} conducted an ablation study by removing the gating mechanism and other nonlinear components. The results showed that comparable accuracy, and in some cases slightly improved accuracy, could be achieved without gating. Motivated by this observation, subsequent models, such as MACE~\cite{MACE}, also avoid introducing additional nonlinear transformations after aggregation. \textbf{In fact, as the scale of the dataset increases, the fitting capacity of purely linear models becomes limited. Therefore, nonlinear modules are necessary. However, improved accuracy should not be obtained at the cost of a substantial loss in extrapolation capability or by introducing unconstrained modules. Instead, appropriate nonlinearities should be incorporated while preserving the model's extrapolation performance}. Accordingly, the design principle of TACE is not to perform module ablations solely based on in-distribution training accuracy. Instead, we evaluate each candidate module on the 3BPA benchmark at 1200K. Any module that increases the force RMSE beyond 65 meV/\AA is excluded from the final architecture.

\paragraph{Differences Between Edge/Atomic Cluster Expansion.}
  Here, we discuss the differences between ECE and ACE. In a local frame, edge features ($m > 0$) cannot be aggregated arbitrarily. Consequently, we cannot employ a bilinear operation analogous to that used in Allegro~\cite{Allegro} (global SO(3) frame) to introduce an explicit aggregation step before ECE (local SO(2) frame). Compared with ACE, the many-body information in ECE is represented more implicitly. Moreover, ECE is generally not applied in the first layer, since the corresponding edge features have not yet aggregated information from all neighboring atoms. The role of ECE is to increase the effective body order of the model while explicitly modeling the relationship between the source and target atoms. When combined with appropriate nonlinearities, ECE can build nonlinear messages. The incorporation of nonlinearities in ECE follows the same procedure as in ACE and details can be found in Section~\ref{ace}.

\subsection{Angular Resolution and Body Order}
\label{sec:mfoldandkbody}
  In this subsection, we analyze why Edge Cluster Expansion can increase angular resolution and body order, namely, whether it is complete in this sense. Specifically, we study the ability of these modules to resolve $M$-fold rotationally symmetric structures and whether they can increase the effective body order of the representation, and, if so, through what mechanism this increase is achieved.
  Joshi et al. identify two key factors that are critical to the a~\cite{mfold} identifies two key factors that are critical to the accuracy of MLIPs: the angular resolution $L_{max}$ (or, more precisely, $m_{max}$ in this work) and the body order ($\nu$). In particular, they validate this conclusion through two experiments. We redesign these experiments and use them to illustrate the importance of ECE from the perspectives of global SO(2) frame and global SO(3) frame.
  
\paragraph{Global SO(2) Frame.}
  The first experiment compares models with different values of $m_{\max}$ in distinguishing 2-dimension $M$-fold rotationally symmetric structures with different global orientations~\cite{mfold}. Since the structures considered here are planar, we describe them using the SO(2) group rather than the SO(3) group adopted. Moreover, we provide an explicit mathematical derivation that offers a more rigorous treatment, replacing the less formal discussion in the original work. Since the two structures differ only in their global orientations and are otherwise symmetric, we denote the orientation by $\phi$ and ignore the radial component. We directly use the edge vector $\hat{\mathbf{r}}_{ij}$ to encode angular information: specifically, we compute its azimuthal angle $\theta_{k}$ (Due to the rotational symmetry of the structure, the edges that were originally distinguished by the pair index $ij$ can now be indexed by $k$), which represents the angular positions of the $M$ neighbors surrounding the central atom and construct the corresponding complex SO(2) angular features:
  \begin{equation}
  \mathbf{x}_{m}^{k}(\phi)
  =
  e^{\mathrm{i}m\theta_k(\phi)}
  =
  e^{\mathrm{i}m\left(\phi+\frac{2\pi k}{L}\right)},
  \qquad
  k=0,\ldots,M-1.
  \end{equation}
  Assume that the model constructs the final representation of the central atom by directly aggregating information from its neighboring atoms:
  \begin{equation}
  A_m(\phi)
  =
  \sum_{k=0}^{M-1} e^{\mathrm{i}m\theta_k(\phi)}
  =
  \begin{cases}
  M e^{\mathrm{i}m\phi},
  & m\equiv 0\pmod M,\\
  0,
  & \text{otherwise}.
  \end{cases}
  \end{equation}
  It follows immediately that if the input angular resolution satisfies $m_{\max}<M$, then the aggregated information at center atom is independent of the global orientation $\phi$. Moreover, applying an SO(2) Atomic Cluster Expansion after the aggregation cannot resolve this issue, because the relevant angular information has already vanished during the aggregation:
  \begin{equation}
  A_{m_1}(\phi)A_{m_2}(\phi)=0,
  \qquad
  m_1,m_2 < M .
  \end{equation}
  In contrast, when the SO(2) Edge Cluster Expansion product is applied before aggregation, complex multiplication combines angular frequencies and thereby generates higher angular orders. As a result, the model can obtain representations that depend on the global orientation:
  \begin{equation}
  E_{m_1}(\phi)E_{m_2}(\phi)\neq0,
  \qquad
  m_1+m_2 = M .
  \end{equation}
  It is worth noting that although we do not explicitly include learnable weights in the notation here, the conclusion remains unchanged when such weights are used. Since all features are represented in a global SO(2) frame, applying SO(2) Linear before or after aggregation is equivalent. Accordingly, this experiment exposes a potential limitation of SO(2) ACE in this setting, while also highlighting the importance of SO(2) ECE, even though body order is not involved in this case.

\paragraph{Global SO(3) Frame.}
  Using the 2-body, 3-body, and 4-body counterexamples from Ref.~\cite{Incompleteness}, we illustrate how ECE progressively constructs higher-body-order features. Each $k$-body counterexample consists of two geometric graphs that are indistinguishable under $k$-body scalarization. In general, without an explicit many-body expansion, a single message-passing layer is equivalent to a $2$-body scalarization and therefore cannot distinguish the $2$-body counterexample. Increasing the correlation order in the many-body expansion progressively improves the completeness of the model. For an ACE-based architecture, the computation follows the pipeline:
  \begin{equation*}
  \begin{aligned}
  \mathrm{Convolution}
  &\rightarrow
  \mathrm{Aggregation}
  \rightarrow
  \mathrm{ACE}
  \rightarrow
  \mathrm{Node\ Readout}.
  \end{aligned}
  \end{equation*}
  For the edge-based setting, however, additional design is required to properly test the effective body order of the model. Specifically, we use the following pipeline:
  \begin{flalign*}
  &\mathrm{Convolution}
  \rightarrow
  \mathrm{Aggregation}
  \rightarrow
  \mathrm{Gather}
  \rightarrow
  \mathrm{Rotate}
  \rightarrow
  &&\\
  &\mathrm{ECE}
  \rightarrow
  \mathrm{SO(2) Linear}
  \rightarrow
  \mathrm{Rotate\ Back}
  \rightarrow
  \mathrm{Edge\ Readout}.
  &&
  \end{flalign*}
  Using ECE also requires first aggregating the two-body scalarized information. Therefore, the pipeline up to the aggregation step is the same as in the ACE setting. However, since ECE is performed on edges and under an $\mathrm{SO}(2)$ representation, additional gather, rotate, and rotate-back operations are required. It is important to note that once an SO(2) Linear operation is applied in this rotated frame, higher-order components with $l>0,m=0$ can be mixed into the scalar component with $l=0,m=0$. This is equivalent to introducing an interaction with the edge $ij$, thereby increasing the effective body order by one. As a result, the model becomes capable of distinguishing the $2$-body counterexample. However, it should be emphasized that stacking multiple SO(2) Linear layers still does not further increase the effective body order, and therefore the model remains limited to distinguishing the $2$-body counterexample. The reason why SO(2) Linear is necessary in this pipeline is that channel-wise ECE alone, although capable of increasing the body order, cannot transfer the resulting higher-order information into the scalar (l=0, m=0). Due to the block-diagonal structure of the Wigner-D matrices, information in the $l>0,m=0$ components cannot flow into the $l=0,m=0$ component without an SO(2) Linear mixing step. In this case, the operation only improves the angular resolution, as in the previous experiment, while the higher-order scalarized information is not incorporated into the final readout function. This motivates the pipeline design described above. Therefore, since the SO(2) Linear operation introduces exactly one additional effective body order, the results in Table~\ref{tab:body_order_experiments} should be interpreted carefully. Finally, in our experiments, the SO(2) Linear operation is implemented as a special case that permutes the components $l>0,m=0$ with scalar component $l=0,m=0$. As shown in Table~\ref{tab:body_order_experiments}, ECE indeed increases the effective body order, thereby demonstrating its effectiveness and completeness in distinguishing higher-body-order geometric structures.
  \begin{table}[t]
      \centering
      \caption{
      Results of the body-order experiments from~\cite{Incompleteness}. Each $k$-body counterexample consists of two geometric graphs that cannot be distinguished by $k$-body scalarization. Architectures that successfully distinguish the two graphs achieve $100\%$ accuracy and are marked in {\setlength{\fboxsep}{0pt}\colorbox{green!10}{\strut green}}; otherwise, they achieve $50\%$ accuracy and are marked in {\setlength{\fboxsep}{0pt}\colorbox{red!10}{\strut red}}.
      }
      \scalebox{0.65}
      {
      \begin{tabular}{lcccc}
      \toprule[1.2pt]
      & & \multicolumn{3}{c}{Counterexamples from~\cite{Incompleteness}} \\
      \cmidrule(lr){3-5}
      & & 2-body & 3-body & 4-body \\
      Architecture & & & {\textcolor{gray}{(Fig. 1(b))}} & {\textcolor{gray}{(Fig. 2(f))}} \\
      \midrule[1.2pt]

      Edge Cluster Expansion  & & & & \\
      Without SO(2) Linear  & & & & \\
      \quad no-rotate & & \cellcolor{red!10} \quad \quad 50.0 \quad \quad & \cellcolor{red!10} \quad \quad 50.0 \quad \quad & \cellcolor{red!10} \quad \quad 50.0 \quad \quad \\
      \quad rotate + $\nu=1$ & & \cellcolor{red!10} \quad \quad 50.0 \quad \quad & \cellcolor{red!10} \quad \quad 50.0 \quad \quad & \cellcolor{red!10} \quad \quad 50.0 \quad \quad \\
      \quad rotate + $\nu=2$ & & \cellcolor{red!10} \quad \quad 50.0 \quad \quad & \cellcolor{red!10} \quad \quad 50.0 \quad \quad & \cellcolor{red!10} \quad \quad 50.0 \quad \quad \\
      \quad rotate + $\nu=3$ & & \cellcolor{red!10} \quad \quad 50.0 \quad \quad & \cellcolor{red!10} \quad \quad 50.0 \quad \quad & \cellcolor{red!10} \quad \quad 50.0 \quad \quad \\
      \midrule

      One SO(2) Linear  & & & & \\
      Without Edge Cluster Expansion & & & & \\
      \quad no-rotate & & \cellcolor{red!10} \quad \quad 50.0 \quad \quad & \cellcolor{red!10} \quad \quad 50.0 \quad \quad & \cellcolor{red!10} \quad \quad 50.0 \quad \quad \\
      \quad rotate + $\nu=1$ & & \cellcolor{green!10} \quad \quad 100.0 \quad \quad & \cellcolor{red!10} \quad \quad 50.0 \quad \quad & \cellcolor{red!10} \quad \quad 50.0 \quad \quad \\
      \quad rotate + $\nu=2$ & & \cellcolor{green!10} \quad \quad 100.0 \quad \quad & \cellcolor{red!10} \quad \quad 50.0 \quad \quad & \cellcolor{red!10} \quad \quad 50.0 \quad \quad \\
      \quad rotate + $\nu=3$ & & \cellcolor{green!10} \quad \quad 100.0 \quad \quad & \cellcolor{red!10} \quad \quad 50.0 \quad \quad & \cellcolor{red!10} \quad \quad 50.0 \quad \quad \\
      \midrule

      Edge Cluster Expansion  & & & & \\
      With One SO(2) Linear  & & & & \\
      \quad no-rotate & & \cellcolor{red!10} \quad \quad 50.0 \quad \quad & \cellcolor{red!10} \quad \quad 50.0 \quad \quad & \cellcolor{red!10} \quad \quad 50.0 \quad \quad \\
      \quad rotate + $\nu=1$ & & \cellcolor{green!10} \quad \quad 100.0 \quad \quad & \cellcolor{red!10} \quad \quad 50.0 \quad \quad & \cellcolor{red!10} \quad \quad 50.0 \quad \quad \\
      \quad rotate + $\nu=2$ & & \cellcolor{green!10} \quad \quad 100.0 \quad \quad & \cellcolor{green!10} \quad \quad 100.0 \quad \quad & \cellcolor{red!10} \quad \quad 50.0 \quad \quad \\
      \quad rotate + $\nu=3$ & & \cellcolor{green!10} \quad \quad 100.0 \quad \quad & \cellcolor{green!10} \quad \quad 100.0 \quad \quad & \cellcolor{green!10} \quad \quad 100.0 \quad \quad \\
      
      \bottomrule[1.2pt]
      \end{tabular}
      }    
      \label{tab:body_order_experiments}
  \end{table}

\subsection{Radial Rotary Complex Attention}
  \label{RRA}
  We argue that geometric graphs are inherently weighted by radial information. Discarding this information and relying solely on simple QK inner products or contractions with attention vectors may impair the model's extrapolation capability. Ideally, attention should explicitly depend on both the similarity between nodes and their relative distances. Drawing inspiration from Rotary Position Embeddings (RoPE)~\cite{RoPE}, we propose Radial Rotary Complex Attention (RRA). Given the complex-valued nature of SO(2) representations, we define the QK similarity as the real part (0e) of the complex inner product. We discard the imaginary component because it transforms as $0o$. We further introduce a phase term and a bias term, which directly modulate the QK inner product as a scale and a shift, respectively. \textbf{Notably, although the scale term is, in principle, allowed to be complex-valued, using a complex-valued phase term restricts the model to SO(2) equivariance in the local frame and breaks O(2) equivariance. In contrast to the weights in SO(2) Linear layers, we observe that the use of a complex-valued phase term does not affect model extrapolation and accuracy}. We denote SO(2) features in local frame as $\mathbf{x}^{ij}_{j,m,h,c}$ and $\mathbf{x}^{ij}_{i,m,h,c}$, respectively, where $m$ is the SO(2) order, $h$ denotes the attention head and $c$ denotes the channel within each head. \textbf{The complex query, key, and value features are obtained using independent uvSO2Linear with weight type \texttt{w1}, respectively. Notably, the phase term and bias shift are deliberately designed as zero-initialized linear mappings that depend only on radial basis. We intentionally exclude element information from these two terms, so that they encode purely distance-dependent effects. Element-dependent feature similarity is instead captured through the complex inner product.}
  \begin{equation}
  \begin{aligned}
  \mathbf{Q}_{m,h,c}^{ij} &= \text{uvSO2Linear}
  \left(\mathbf{x}^{ij}_{i,m,h,c}\right), \\
  \mathbf{K}_{m,h,c}^{ij} &= \text{uvSO2Linear}
  \left(\mathbf{x}^{ij}_{j,m,h,c}\right), \\
  \mathbf{V}_{m,h,c}^{ij} &= \text{uvSO2Linear}
  \left(\mathbf{x}^{ij}_{i,m,h,c}, \mathbf{x}^{ij}_{j,m,h,c}\right).
  \end{aligned}
  \end{equation}
  The attention logit is defined as:
  \begin{equation}
  s^{ij}_{h}
  =
  b^{ij}_{h}
  +
  \frac{\tau_h}{\sqrt{D}}
  \sum_{m=0}^{m_{\max}}
  \sum_{c}
  \operatorname{Re}
  \left[
  e^{\mathrm{i}\theta_{h}}
  \,
  \overline{\mathbf{Q}}^{ij}_{m,h,c}
  \,
  {\mathbf{K}}^{ij}_{m,h,c}
  \right].
  \end{equation}
  In particular, we define radial phase $\theta_{h}$, radial bias $b_{h}$ and and temperature $\tau_h$ as:
  \begin{equation}
  \theta_{m,h}(r) = 0 \, \text{or} \, m \pi \cdot \tanh
  \left(b_{h} + \sum_{n}W_{h}^{n}j_0^n(r)\right),
  \end{equation}
  \begin{equation}
  b_{h}(r) = b_{h} + \sum_{n}W_{h}^{n}j_0^n(r),
  \end{equation}
  \begin{equation}
  \tau_h
  = \tau_{\min} + \frac{\tau_{\max}-\tau_{\min}}{1+e^{-w_h}}.
  \end{equation}
  where W and b represent the learnable weight and bias, respectively, n represent the number of radial basis. Here, $\operatorname{Re}(\cdot)$ denotes the real-part operator. The \textbf{optional} temperature term $\tau_{h}$ controls the sensitivity of the softmax function across different attention heads. In practice, we typically set $\tau_{\min}=0.25$ and $\tau_{\max}=4.0$. The attention weight for each head is then computed over all incoming edges of node $i$. Meanwhile, an envelope function is incorporated to imporve the discontinuity issue reported in EquiformerV3~\cite{eqv3},
  \begin{equation}
  \alpha^{ij}_{h}
  =
  \frac{
  \exp
  \left(
  s^{ij}_{h}
  \right)\cdot f_{\text{cut}}(r_{ij})
  }{
  \sum_{k \in \mathcal{N}(i)}
  \exp
  \left(
  s^{ik}_{h} 
  \right)\cdot f_{\text{cut}}(r_{ik})
  }.
  \end{equation}
  To examine the effect of RRA, we compare different attention mechanisms, with the results summarized in Table~\ref{tab:RRCA}. The best results are highlighted in \colorbox{green!10}{green}. Notably, even when using \texttt{uvSO2Linear}, which may compromise extrapolation performance, our model still achieves strong results. This suggests that the strong inductive bias introduced by RRA can effectively compensate for this limitation, thereby supporting our claim.

\subsection{Coupled Atomic Cluster Expansion.}
  \label{ace}
  We still adopt the original coupled ACE implementation used in the Cartesian TACE. Specifically, ``coupled'' means that, after the tensor product, element-dependent weights are applied to couple information across all channels. In practice, using only $48$-$64$ channels is sufficient to achieve an accuracy comparable to that obtained by the uncoupled implementation with $128$-$256$ channels. However, its limitations are also evident. First, the element-dependent parameterization introduces a large number of weights. Second, when the number of channels exceeds $64$, the coupled formulation leads to unnecessary computational and memory overhead. Therefore, we improve the ACE implementation mainly in two aspects. First, we make coupled ACE feasible when increasing the channel dimension. Second, we introduce additional nonlinearities without destroying the resulting many-body structure.

  \paragraph{From Dense Linear to Group Linear.}
  Unlike standard sparse Mixture-of-Experts models~\cite{moe}, where each selected expert is still a linear transformation over all input channels and using top-$k$ selection, our goal is to keep computation feasible as the number of channels increases. Therefore, such a MoE formulation does not directly address our problem. Instead, we uniformly partition the input and output channels into $E$ element-aware expert groups, so that each expert only operates on a subset of channels. To compensate for the reduced cross-channel communication, we additionally introduce an element-agnostic shared expert that couples all channels,
  \begin{equation}
  \underbrace{
  \begin{bmatrix}
  \mathbf{W}^{(z)}_1 & \mathbf{0} & \cdots & \mathbf{0} \\
  \mathbf{0} & \mathbf{W}^{(z)}_2 & \cdots & \mathbf{0} \\
  \vdots & \vdots & \ddots & \vdots \\
  \mathbf{0} & \mathbf{0} & \cdots & \mathbf{W}^{(z)}_E
  \end{bmatrix}
  }_{\text{aware group experts}}
  +
  \underbrace{
  \mathbf{W}_{\mathrm{shared}}
  }_{\text{agnostic shared expert}}.
  \end{equation}
  Here, $\mathbf{W}$ denotes learnable weights, and $z$ denotes the element type on which the group-expert weights depend.

  \paragraph{GLU-style implementation and Nonlinear Coefficients.}
  To introduce additional nonlinearity into the product basis block without disrupting its many-body structure, we adopt an EquflashV2-inspired strategy, in which scalar 0e features are used to parameterize the tensor product weights~\cite{glu,equflashv2}. An linear projection first generates two asymmetric atomic bases together with scalar, feature-dependent ACE coefficients:
  \begin{equation}
  \operatorname{Linear}_{\mathrm{up}}(\mathbf{A}_{i})
  =
  \mathrm{A}_{i}^{0e}
  \oplus
  \mathbf{A}_{i}^{\mathrm{node}}
  \oplus
  \mathbf{A}_{i}^{\mathrm{base}},
  \end{equation}
  where both $\mathbf{A}_{i}^{\mathrm{node}}$ and $\mathbf{A}_{i}^{\mathrm{base}}$ contain all retained irreps, including their $0e$ components. The first-order correlation is defined directly by the node features, while the second-order correlation is obtained by coupling them with the base
  features augmented by a constant scalar:
  \begin{equation}
  \mathbf{B}_{i}^{(1)}
  =
  \mathbf{A}_{i}^{\mathrm{node}}
  \end{equation}
  \begin{equation}
  \mathbf{B}_{i}^{(2)}
  =
  \underbrace{
  \mathbf{A}_{i}^{\mathrm{node}}
  \otimes_{\mathrm{SO(3)}}
  \left(
  \mathbf{1}_{0e}
  \oplus
  \mathbf{A}_{i}^{\mathrm{base}}
  \right)
  }_{
  \text{modulated by }
  \operatorname{SiLU}\!\left(\mathrm{A}_{i}^{0e}\right)
  }.
  \end{equation}
  The constant scalar $\mathbf{1}_{0e}$ and the base features are used to approximately produce second- and third-order correlations, respectively, despite involving only a single tensor-product operation. The resulting correlation features are then concatenated by $
  \mathbf{B}_{i} = \mathbf{B}_{i}^{(1)} \oplus \mathbf{B}_{i}^{(2)}$. Finally, the grouped and shared experts are combined using
  %
  \begin{equation}
  \frac{
  \operatorname{Linear}_{\mathrm{group}}(\mathbf{B}_{i})
  +
  \operatorname{Linear}_{\mathrm{shared}}(\mathbf{B}_{i})
  }{\sqrt{2}}.
  \end{equation}

\subsection{Residual Network and Layer Norm}
  In equivariant networks, an important detail is that the first layer in the message-passing process differs from subsequent layers. During the first edge-level tensor-product operation, the tensor product is typically formed between 0e features (such as element embedding) and spherical harmonics. Consequently, prior to the first message-passing step, only scalar information is available. Under such circumstances, directly treating the 0e featrues as residual connections~\cite{resnet} is generally unfavorable. We further observed that, in the absence of the LayerNorm layer, employing two residual connections independently in both the interaction module ($\mathbf{A}$ basis) and the product module ($\mathbf{B}$ basis) does not yield optimal performance in our framework. Moreover, this design becomes more sensitive to the number of layers and is more prone to training instability. Instead, superior accuracy is consistently achieved when only a single residual pathway is used. In particular, both the $\mathbf{A} \rightarrow \mathbf{B}$ and $\mathbf{B} \rightarrow \mathbf{B}$ connection schemes outperform the $\mathbf{B} \rightarrow \mathbf{A} \rightarrow \mathbf{B}$ design, regardless of whether the identity branch is implemented through direct addition or (element-dependent) linear transformations. We also note that recently proposed residual-network designs, such as Attention Residuals (AttnRes)~\cite{AttnRes}, have been shown to further improve model accuracy. However, since the depth of current uMLIPs remain substantially smaller than that of modern LLMs, the benefits brought by such residual formulations are comparatively limited. Therefore, although these approaches may provide additional accuracy gains, we do not adopt them in the present work in order to reduce computational overhead. We also observed that, compared with non-ACE architectures, ACE-based models are generally more sensitive to outliers and tend to exhibit greater training instability when multiple layers are stacked. To address this issue, we adopt the \texttt{EquivariantMergeLayerNorm} implementation introduced in EquiformerV3~\cite{eqv3}. In our view, this normalization strategy currently provides one of the most effective solutions for equivariant architectures, as it preserves the scale differences across different irreps while improving training stability. Following the same considerations applied to the residual design, we omit LayerNorm in the first layer and do not apply it after the final layer. This is because LayerNorm is primarily used to stabilize training, whereas an additional normalization at the output stage of MLIPs may limit the model's expressive accuracy. Hence, unless the training process is highly unstable, introducing LayerNorm after the final layer is unnecessary.

\subsection{Effects of Optimizers on Model Extrapolation}
  Recently, emerging optimizers such as Muon~\cite{Muon} and SOAP~\cite{SOAP} have attracted increasing attention as potential alternatives to conventional optimizers such as Adam~\cite{Adam} and AdamW~\cite{AdamW}. For example, Harari et al.~\cite{SOAP_Muon} systematically investigated SOAP and Muon and showed that both optimizers, whether used individually or in combination, consistently achieved higher accuracy. Similarly, DPA-4~\cite{DeePMD-kit, DPA4} employed the Muon optimizer and demonstrated strong performance on the MPtrj dataset. These emerging optimizers often provide improved in-domain accuracy and faster convergence. Nevertheless, we emphasize that these optimizers should be used with caution, as both Muon and SOAP may compromise extrapolation performance to some extent. In terms of extrapolation capability, Adam performs best, followed by SOAP and then Muon.

\subsection{Matbench Discovery}
  Matbench Discovery~\cite{Matbench} has become a widely used benchmark for evaluating foundation models in materials science. In particular, OMat24, sAlex, and MPtrj~\cite{CHGNet,Alex,OMat24} are among the most advanced and widely used materials datasets currently available. We therefore train our model on these datasets to demonstrate its representational capacity. As shown in Table~\ref{tab:matbench}, our model achieves state-of-the-art (SOTA) performance among compared architectures. Notably, among the three most important metrics, F1 score for crystal stability prediction, thermal conductivity prediction, and structure optimization, our model achieved the best performance on two tasks and ranked second on the remaining one, resulting in the highest overall score.

  \begin{table}[!h]
    \caption{Top models on MatBench Discovery~\cite{Matbench}. For models with the same architecture, we retain only the variant with the highest accuracy. CPS is a weighted combination of the F1 score, thermal-conductivity metric ($\kappa$SRME), and RMSD, and thus represents the model with the strongest overall performance. Full results are provided in Appendix Table~\ref{tab:full_matbench}.}
    \label{tab:matbench}
    \begin{center}
      \scalebox{0.65}
      {
      \begin{tabular}{lcccccr}
        \toprule
        Model & CPS & F1($\uparrow$) & $\kappa$SRME ($\downarrow$)& RMSD& Params \\
        \midrule
        \rowcolor{green!10}
        TECE-OAM-RRA-1.0 & \textbf{0.908} & 0.929 & \textbf{0.093} & \textbf{0.058} & 222M \\
        EquFlashV2 & 0.907 & 0.929 & 0.094 & \textbf{0.058} & 44.9M \\
        EquiformerV3+DeNS-OAM & 0.902 & \textbf{0.931} & 0.118 & 0.060 & 30.3M \\
        GRACE-3L-OAM-L & 0.900 & 0.925 & 0.121 & \textbf{0.058} & 42.1M \\
        PET-OAM-XL & 0.898 & 0.924 & 0.119 & 0.060 & 730M \\
        TACE-OAM-L & 0.889 & 0.910 & 0.126 & 0.061 & 82.9M \\
        eSEN-30M-OAM & 0.888 & 0.925 & 0.170 & 0.061 & 30.2M \\
        EquFlash & 0.888 & 0.919 & 0.158 & 0.060 & 28.7M \\
        Nequip-OAM-XL & 0.886 & 0.906 & 0.125 & 0.063 & 32.1M \\
        MatRIS-10M-OAM & 0.877 & 0.921 & 0.218 & 0.060 & 10.4M \\
        SevenNet-Omni-i12 & 0.873 & 0.906 & 0.192 & 0.062 & 54.9M \\
        GRACE-2L-OAM-L & 0.865 & 0.883 & 0.169 & 0.064 & 26.4M \\
        ORB v3 & 0.860 & 0.905 & 0.210 & 0.075 & 25.5M \\
        DPA-4.0.1-Pro-MPtrj & 0.840 & 0.857 & 0.211 & 0.069 & 22.8M \\
        Allegro-OAM-L & 0.840 & 0.895 & 0.319 & 0.065 & 9.7M \\
        \bottomrule
      \end{tabular}
      }
    \end{center}
    \vskip -0.1in
  \end{table}

\subsection{Model Speed}
  We are pleased to see the emergence of systematic evaluations of both the accuracy and computational efficiency of uMLIPs, as these benchmarks provide valuable guidance for the community when selecting models across a wide range of model sizes~\cite{md_benchmark}. Nevertheless, we would like to emphasize an important point. Improving both model extrapolation and fitting accuracy remains essential~(see Section~\ref{par:extrapolation}). A more expressive architecture can achieve accuracy with a smaller and faster model, while its highest-accuracy variant can also serve as an effective teacher for knowledge distillation. In practice, smaller variants of SOTA architectures can therefore offer a better combination of accuracy and speed, since increasing model size often yields only marginal improvements on accuracy-oriented benchmarks. 
  Existing speed benchmarks do not always evaluate models under their optimal configurations. In some cases, appropriate acceleration techniques may not be enabled because of limited familiarity with the model implementation. To provide a fairer and more representative comparison, we benchmarked the top three models~\cite{eqv3, equflashv2} on Matbench Discovery using the best available configuration for each model and compared their practical runtime performance. As shown in Table~\ref{tab:speed}, the top three models are ranked in terms of computational speed and memory efficiency as follows: EquflashV2 $>$ TECE-OAM-RRA-1.0 $>$ EquiformerV3+DeNS-OAM. This result is consistent with our expectations. As discussed previously, architectures based on SO(2) should employ \texttt{uuSO2Linear} to enable operator fusion. In comparison, EquflashV2 adopts more efficient SO(3) edge-fused operators, resulting in improved computational performance. Nevertheless, SO(2) architectures may still offer a higher performance ceiling and more favorable scaling behavior. We therefore look forward to further developments in efficient SO(2) kernel.
  \begin{table}[ht!]
    \caption{This table reports the molecular dynamics simulation speed (float32), measured in ps/day ($\uparrow$), for top-3 models (CPS) on Matbench Discovery using ASE~\cite{ASE} as the simulation engine. The comparison is performed using a single NVIDIA A800 80GB PCIe GPU.}
    \label{tab:speed}
    \begin{center}
      \begin{small}
        \scalebox{0.75}
        {
        \begin{tabular}{lccr}
        \toprule
        Atoms & TECE-OAM-RRA-1.0 & EquflashV2 & EquiformerV3+DeNS-OAM\\
        \midrule      
        384  & 144 & 285 & 92 \\
        768  & 77  & 255 & OOM  \\
        960  & 62  & 240 & OOM  \\
        1152 & OOM & 210 & OOM  \\
        3264 & OOM & 83  & OOM  \\
        3456 & OOM & OOM & OOM  \\
        \bottomrule
        \end{tabular}
        }
      \end{small}
    \end{center}
  \end{table}

\section{Discussion, Limitations, and Outlook}

\subsection{Diatomic Benchmark}
  We also observed several interesting phenomena in the diatomic benchmark. The widely used OAM-family datasets exhibit discontinuities in their potential energy surfaces when O or F occurs with certain transition-metal elements, due to the use of different Hubbard U treatments~\cite{without_U}. In addition, OrbV3~\cite{OrbV3} also reported that the rattle subset of OMat24 can adversely affect performance on diatomic benchmarks. Indeed, the top 3 models on Matbench Discovery, including TECE-OAM-RRA-1.0, EquFlashV2, and EquiformerV3+DeNS-OAM consistently exhibit relatively poor performance on the diatomic benchmark. By contrast, models with lower fitting capacity, such as TACE-OAM-L, NequIP-OAM-XL, and MACE-MPA-0, tend to perform better on this benchmark. We attribute this behavior primarily to limitations in the training data: as model capacity and fitting accuracy increase, the model becomes more likely to fit dataset-specific artifacts and unphysical features, thereby reducing its robustness in regimes such as diatomic configurations.

  In summary, we systematically review, consolidate, and extend SO(2) theory for machine learning interatomic potentials, while validating several implementation details that are important in practice. We further introduce Cartesian direct and recursive Wigner-D constructions, Edge Cluster Expansion (ECE), and Radial Rotary Complex Attention (RRA), and release a SOTA materials foundation model built upon these components. Our results suggest that the proposed design choices provide complementary benefits rather than a universally optimal configuration. When pursuing the highest accuracy in large-scale pretraining, ECE with edge nonlinearities and RRA provides the most effective strategy. When memory efficiency is the primary concern, \texttt{uuSO2Linear} is generally preferable because of its stronger inductive bias, expressive capability, and compatibility with operator fusion. From an architectural perspective, an important direction for future work is operator fusion for single-layer SO(2) Linear operators. During the course of this work, we also confirmed several limitations in the existing datasets, including data redundancy, inconsistent DFT parameter settings, and inappropriate treatments of Hubbard U corrections. For example, TACE-OAM-L, an intermediate model developed during the construction of our SO(2) model family, was trained on only approximately random 20\% of the training data due to a code bug. Despite this substantial reduction in training data, it still exhibited remarkably strong performance.

\section*{Author contribution}
W.X. and P.H. conceived the project and guided the research. Z.X. prepared code, equations, figures and tables. All authors edited and revised the manuscript.

\section*{Acknowledgment}
This work was supported by the National Natural Science Foundation of China NSFC (22433004 and 22403064), the open research fund of Key Laboratory of Precision and Intelligent Chemistry, and ShanghaiTech University. We are also grateful for the computing time provided by the the HPC Platform of ShanghaiTech University.

\bibliography{references}

\begin{thebibliography}{51}
\providecommand{\natexlab}[1]{#1}
\providecommand{\url}[1]{\texttt{#1}}
\expandafter\ifx\csname urlstyle\endcsname\relax
  \providecommand{\doi}[1]{doi: #1}\else
  \providecommand{\doi}{doi: \begingroup \urlstyle{rm}\Url}\fi

\bibitem[Barros-Luque et~al.(2026)Barros-Luque, Shuaibi, Fu, Wood, Dzamba, Gao, Rizvi, Uyttendaele, Zitnick, and Ulissi]{OMat24}
Barros-Luque, L., Shuaibi, M., Fu, X., Wood, B.~M., Dzamba, M., Gao, M., Rizvi, A., Uyttendaele, M., Zitnick, C.~L., and Ulissi, Z.~W.
\newblock The open materials 2024 (omat24) inorganic materials dataset and models.
\newblock \emph{Nature Computational Science}, pp.\  1--11, 2026.

\bibitem[Batatia et~al.(2022)Batatia, Kovacs, Simm, Ortner, and Csanyi]{MACE}
Batatia, I., Kovacs, D.~P., Simm, G., Ortner, C., and Csanyi, G.
\newblock Mace: Higher order equivariant message passing neural networks for fast and accurate force fields.
\newblock In Koyejo, S., Mohamed, S., Agarwal, A., Belgrave, D., Cho, K., and Oh, A. (eds.), \emph{Advances in Neural Information Processing Systems}, volume~35, pp.\  11423--11436. Curran Associates, Inc., 2022.

\bibitem[Batatia et~al.(2025)Batatia, Batzner, Kov{\'a}cs, Musaelian, Simm, Drautz, Ortner, Kozinsky, and Cs{\'a}nyi]{BotNet}
Batatia, I., Batzner, S., Kov{\'a}cs, D.~P., Musaelian, A., Simm, G.~N., Drautz, R., Ortner, C., Kozinsky, B., and Cs{\'a}nyi, G.
\newblock The design space of e (3)-equivariant atom-centred interatomic potentials.
\newblock \emph{Nature Machine Intelligence}, 7\penalty0 (1):\penalty0 56--67, 2025.

\bibitem[Batzner et~al.(2022)Batzner, Musaelian, Sun, Geiger, Mailoa, Kornbluth, Molinari, Smidt, and Kozinsky]{NequIP}
Batzner, S., Musaelian, A., Sun, L., Geiger, M., Mailoa, J.~P., Kornbluth, M., Molinari, N., Smidt, T.~E., and Kozinsky, B.
\newblock E (3)-equivariant graph neural networks for data-efficient and accurate interatomic potentials.
\newblock \emph{Nature communications}, 13\penalty0 (1):\penalty0 2453, 2022.

\bibitem[Bharadwaj et~al.(2025)Bharadwaj, Glover, Bulu{\c{c}}, and Demmel]{oeq}
Bharadwaj, V., Glover, A., Bulu{\c{c}}, A., and Demmel, J.
\newblock An efficient sparse kernel generator for o (3)-equivariant deep networks.
\newblock In \emph{2025 Proceedings of the Conference on Applied and Computational Discrete Algorithms (ACDA)}, pp.\  32--46. SIAM, 2025.

\bibitem[Bigi et~al.(2026)Bigi, Pegolo, Mazitov, and Ceriotti]{PET}
Bigi, F., Pegolo, P., Mazitov, A., and Ceriotti, M.
\newblock Pushing the limits of unconstrained machine-learned interatomic potentials, 2026.
\newblock URL \url{https://arxiv.org/abs/2601.16195}.

\bibitem[Dauphin et~al.(2017)Dauphin, Fan, Auli, and Grangier]{glu}
Dauphin, Y.~N., Fan, A., Auli, M., and Grangier, D.
\newblock Language modeling with gated convolutional networks.
\newblock In Precup, D. and Teh, Y.~W. (eds.), \emph{Proceedings of the 34th International Conference on Machine Learning}, volume~70 of \emph{Proceedings of Machine Learning Research}, pp.\  933--941. PMLR, 06--11 Aug 2017.
\newblock URL \url{https://proceedings.mlr.press/v70/dauphin17a.html}.

\bibitem[Deng et~al.(2023)Deng, Zhong, Jun, Riebesell, Han, Bartel, and Ceder]{CHGNet}
Deng, B., Zhong, P., Jun, K., Riebesell, J., Han, K., Bartel, C.~J., and Ceder, G.
\newblock Chgnet as a pretrained universal neural network potential for charge-informed atomistic modelling.
\newblock \emph{Nature Machine Intelligence}, 5\penalty0 (9):\penalty0 1031--1041, 2023.

\bibitem[Gawkowski et~al.(2026)Gawkowski, Artrith, Bonfanti, Gangan, Heenen, Kioseoglou, Lončarić, Myneni, Riebesell, Rossi, Rupp, Schmidt, Sharma, Shi, Wadowski, Hörmann, and Kapil]{md_benchmark}
Gawkowski, M.~J., Artrith, N., Bonfanti, S., Gangan, A.~S., Heenen, H.~H., Kioseoglou, J., Lončarić, I., Myneni, H., Riebesell, J., Rossi, M., Rupp, M., Schmidt, J., Sharma, S., Shi, B.~X., Wadowski, A., Hörmann, L., and Kapil, V.
\newblock Dyna-mat: End-to-end benchmarking of foundation machine learning interatomic potentials in finite-temperature ensembles, 2026.
\newblock URL \url{https://arxiv.org/abs/2607.03433}.

\bibitem[Harari et~al.(2026)Harari, Zimmermann, Kulseng, Zichi, Tan, Descoteaux, and Kozinsky]{SOAP_Muon}
Harari, G., Zimmermann, Y., Kulseng, O.~T., Zichi, L., Tan, C.~W., Descoteaux, M.~L., and Kozinsky, B.
\newblock Beyond adam: Soap and muon for faster, label-efficient training of machine learning interatomic potentials, 2026.
\newblock URL \url{https://arxiv.org/abs/2607.02499}.

\bibitem[He et~al.(2016)He, Zhang, Ren, and Sun]{resnet}
He, K., Zhang, X., Ren, S., and Sun, J.
\newblock Deep residual learning for image recognition.
\newblock In \emph{Proceedings of the IEEE conference on computer vision and pattern recognition}, pp.\  770--778, 2016.

\bibitem[Heyraud et~al.(2026)Heyraud, Weller-Davies, and Tilly]{vstp}
Heyraud, V., Weller-Davies, Z., and Tilly, J.
\newblock Integral formulas for vector signal tensor products, 2026.
\newblock URL \url{https://arxiv.org/abs/2603.08630}.

\bibitem[Huang et~al.(2016)Huang, Sun, Liu, Sedra, and Weinberger]{stochastic_depth}
Huang, G., Sun, Y., Liu, Z., Sedra, D., and Weinberger, K.
\newblock Deep networks with stochastic depth, 2016.
\newblock URL \url{https://arxiv.org/abs/1603.09382}.

\bibitem[Huang et~al.(2026)Huang, Huang, Wang, Du, Wang, Lu, Li, Liu, Jiang, and Zhang]{e2formerv2}
Huang, L., Huang, C., Wang, Z., Du, Y., Wang, C., Lu, H., Li, Y., Liu, X., Jiang, A., and Zhang, J.
\newblock E2former-v2: On-the-fly equivariant attention with linear activation memory.
\newblock \emph{arXiv preprint arXiv:2601.16622}, 2026.

\bibitem[Jiang et~al.(2024)Jiang, Sablayrolles, Roux, Mensch, Savary, Bamford, Chaplot, de~las Casas, Hanna, Bressand, Lengyel, Bour, Lample, Lavaud, Saulnier, Lachaux, Stock, Subramanian, Yang, Antoniak, Scao, Gervet, Lavril, Wang, Lacroix, and Sayed]{moe}
Jiang, A.~Q., Sablayrolles, A., Roux, A., Mensch, A., Savary, B., Bamford, C., Chaplot, D.~S., de~las Casas, D., Hanna, E.~B., Bressand, F., Lengyel, G., Bour, G., Lample, G., Lavaud, L.~R., Saulnier, L., Lachaux, M.-A., Stock, P., Subramanian, S., Yang, S., Antoniak, S., Scao, T.~L., Gervet, T., Lavril, T., Wang, T., Lacroix, T., and Sayed, W.~E.
\newblock Mixtral of experts, 2024.
\newblock URL \url{https://arxiv.org/abs/2401.04088}.

\bibitem[Joshi et~al.(2023)Joshi, Bodnar, Mathis, Cohen, and Lio]{mfold}
Joshi, C.~K., Bodnar, C., Mathis, S.~V., Cohen, T., and Lio, P.
\newblock On the expressive power of geometric graph neural networks.
\newblock 2023.
\newblock URL \url{https://openreview.net/forum?id=Rkxj1GXn9_}.

\bibitem[Kingma \& Ba(2017)Kingma and Ba]{Adam}
Kingma, D.~P. and Ba, J.
\newblock Adam: A method for stochastic optimization, 2017.
\newblock URL \url{https://arxiv.org/abs/1412.6980}.

\bibitem[Kondor et~al.(2018)Kondor, Lin, and Trivedi]{e3nn3}
Kondor, R., Lin, Z., and Trivedi, S.
\newblock Clebsch\textendash gordan nets: a fully fourier space spherical convolutional neural network.
\newblock In Bengio, S., Wallach, H., Larochelle, H., Grauman, K., Cesa-Bianchi, N., and Garnett, R. (eds.), \emph{Advances in Neural Information Processing Systems}, volume~31. Curran Associates, Inc., 2018.

\bibitem[Larsen et~al.(2017)Larsen, Mortensen, Blomqvist, Castelli, Christensen, Du{\l}ak, Friis, Groves, Hammer, Hargus, et~al.]{ASE}
Larsen, A.~H., Mortensen, J.~J., Blomqvist, J., Castelli, I.~E., Christensen, R., Du{\l}ak, M., Friis, J., Groves, M.~N., Hammer, B., Hargus, C., et~al.
\newblock The atomic simulation environment—a python library for working with atoms.
\newblock \emph{Journal of Physics: Condensed Matter}, 29\penalty0 (27):\penalty0 273002, 2017.

\bibitem[Lee et~al.(2025)Lee, Kim, Park, Jeong, Han, Park, and Lee]{equflashv1}
Lee, S.~Y., Kim, H., Park, Y., Jeong, D., Han, S., Park, Y., and Lee, J.~W.
\newblock Flash{TP}: Fused, sparsity-aware tensor product for machine learning interatomic potentials.
\newblock In \emph{Forty-second International Conference on Machine Learning}, 2025.
\newblock URL \url{https://openreview.net/forum?id=wiQe95BPaB}.

\bibitem[Li et~al.(2026{\natexlab{a}})Li, Li, Peng, Xue, Zhang, Zhang, and Wang]{DPA4}
Li, T., Li, W., Peng, A., Xue, J., Zhang, L., Zhang, D., and Wang, H.
\newblock Dpa4: Pushing the accuracy-cost frontier of interatomic potentials with emfa so(2) convolution, 2026{\natexlab{a}}.
\newblock URL \url{https://arxiv.org/abs/2606.02419}.

\bibitem[Li et~al.(2026{\natexlab{b}})Li, Huang, Ding, Wei, Wang, Yang, Wang, Liu, Shi, Jin, Qin, Gerstein, and Zhang]{e2former}
Li, Y., Huang, L., Ding, Z., Wei, X., Wang, C., Yang, H., Wang, Z., Liu, C., Shi, Y., Jin, P., Qin, T., Gerstein, M., and Zhang, J.
\newblock E2former: An efficient and equivariant transformer with linear-scaling tensor products.
\newblock In \emph{The Thirty-ninth Annual Conference on Neural Information Processing Systems}, 2026{\natexlab{b}}.
\newblock URL \url{https://openreview.net/forum?id=ls5L4IMEwt}.

\bibitem[Liao et~al.(2024)Liao, Smidt, Shuaibi, and Das]{DeNS}
Liao, Y.-L., Smidt, T., Shuaibi, M., and Das, A.
\newblock Generalizing denoising to non-equilibrium structures improves equivariant force fields, 2024.
\newblock URL \url{https://arxiv.org/abs/2403.09549}.

\bibitem[Liao et~al.(2026)Liao, Hoffman, Shen, Duval, Norwood, and Smidt]{eqv3}
Liao, Y.-L., Hoffman, A.~J., Shen, S.~C., Duval, A., Norwood, S.~W., and Smidt, T.
\newblock Equiformerv3: Scaling efficient, expressive, and general se (3)-equivariant graph attention transformers.
\newblock \emph{arXiv preprint arXiv:2604.09130}, 2026.

\bibitem[Lin et~al.(2024)Lin, Helwig, Gui, and Ji]{fram_averaging}
Lin, Y., Helwig, J., Gui, S., and Ji, S.
\newblock Equivariance via minimal frame averaging for more symmetries and efficiency.
\newblock In \emph{Forty-first International Conference on Machine Learning}, 2024.
\newblock URL \url{https://openreview.net/forum?id=guFsTBXsov}.

\bibitem[Liu et~al.(2025)Liu, Su, Yao, Jiang, Lai, Du, Qin, Xu, Lu, Yan, Chen, Zheng, Liu, Liu, Yin, He, Zhu, Wang, Wang, Dong, Zhang, Kang, Zhang, Xu, Zhang, Wu, Zhou, and Yang]{Muon}
Liu, J., Su, J., Yao, X., Jiang, Z., Lai, G., Du, Y., Qin, Y., Xu, W., Lu, E., Yan, J., Chen, Y., Zheng, H., Liu, Y., Liu, S., Yin, B., He, W., Zhu, H., Wang, Y., Wang, J., Dong, M., Zhang, Z., Kang, Y., Zhang, H., Xu, X., Zhang, Y., Wu, Y., Zhou, X., and Yang, Z.
\newblock Muon is scalable for llm training, 2025.
\newblock URL \url{https://arxiv.org/abs/2502.16982}.

\bibitem[Loshchilov \& Hutter(2019)Loshchilov and Hutter]{AdamW}
Loshchilov, I. and Hutter, F.
\newblock Decoupled weight decay regularization, 2019.
\newblock URL \url{https://arxiv.org/abs/1711.05101}.

\bibitem[Luo et~al.(2024)Luo, Chen, and Krishnapriyan]{GTP}
Luo, S., Chen, T., and Krishnapriyan, A.~S.
\newblock Enabling efficient equivariant operations in the fourier basis via gaunt tensor products.
\newblock In \emph{The Twelfth International Conference on Learning Representations}, 2024.
\newblock URL \url{https://openreview.net/forum?id=mhyQXJ6JsK}.

\bibitem[Musaelian et~al.(2023)Musaelian, Batzner, Johansson, Sun, Owen, Kornbluth, and Kozinsky]{Allegro}
Musaelian, A., Batzner, S., Johansson, A., Sun, L., Owen, C.~J., Kornbluth, M., and Kozinsky, B.
\newblock Learning local equivariant representations for large-scale atomistic dynamics.
\newblock \emph{Nature Communications}, 14\penalty0 (1):\penalty0 579, 2023.

\bibitem[{NVIDIA}(2024)]{cueq}
{NVIDIA}.
\newblock cuequivariance, 2024.
\newblock URL \url{https://github.com/NVIDIA/cuEquivariance}.

\bibitem[Passaro \& Zitnick(2023)Passaro and Zitnick]{eSCN}
Passaro, S. and Zitnick, C.~L.
\newblock Reducing so (3) convolutions to so (2) for efficient equivariant gnns.
\newblock In \emph{International conference on machine learning}, pp.\  27420--27438. PMLR, 2023.

\bibitem[Paszke et~al.(2019)Paszke, Gross, Massa, Lerer, Bradbury, Chanan, Killeen, Lin, Gimelshein, Antiga, et~al.]{torch}
Paszke, A., Gross, S., Massa, F., Lerer, A., Bradbury, J., Chanan, G., Killeen, T., Lin, Z., Gimelshein, N., Antiga, L., et~al.
\newblock Pytorch: An imperative style, high-performance deep learning library.
\newblock \emph{Advances in neural information processing systems}, 32, 2019.

\bibitem[Pozdnyakov et~al.(2020)Pozdnyakov, Willatt, Bart{\'o}k, Ortner, Cs{\'a}nyi, and Ceriotti]{Incompleteness}
Pozdnyakov, S.~N., Willatt, M.~J., Bart{\'o}k, A.~P., Ortner, C., Cs{\'a}nyi, G., and Ceriotti, M.
\newblock Incompleteness of atomic structure representations.
\newblock \emph{Physical Review Letters}, 125\penalty0 (16):\penalty0 166001, 2020.

\bibitem[Rhodes et~al.(2025)Rhodes, Vandenhaute, {\v{S}}imkus, Gin, Godwin, Duignan, and Neumann]{OrbV3}
Rhodes, B., Vandenhaute, S., {\v{S}}imkus, V., Gin, J., Godwin, J., Duignan, T., and Neumann, M.
\newblock Orb-v3: atomistic simulation at scale.
\newblock \emph{arXiv preprint arXiv:2504.06231}, 2025.

\bibitem[Riebesell et~al.(2025)Riebesell, Goodall, Benner, Chiang, Deng, Ceder, Asta, Lee, Jain, and Persson]{Matbench}
Riebesell, J., Goodall, R.~E., Benner, P., Chiang, Y., Deng, B., Ceder, G., Asta, M., Lee, A.~A., Jain, A., and Persson, K.~A.
\newblock A framework to evaluate machine learning crystal stability predictions.
\newblock \emph{Nature Machine Intelligence}, 7\penalty0 (6):\penalty0 836--847, 2025.

\bibitem[{SamsungDS}(2026)]{equflashv2}
{SamsungDS}.
\newblock Equflashv2, 2026.
\newblock URL \url{https://github.com/SamsungDS/GGNN}.

\bibitem[Schmidt et~al.(2024)Schmidt, Cerqueira, Romero, Loew, J{\"a}ger, Wang, Botti, and Marques]{Alex}
Schmidt, J., Cerqueira, T.~F., Romero, A.~H., Loew, A., J{\"a}ger, F., Wang, H.-C., Botti, S., and Marques, M.~A.
\newblock Improving machine-learning models in materials science through large datasets.
\newblock \emph{Materials Today Physics}, 48:\penalty0 101560, 2024.

\bibitem[Shao et~al.(2025)Shao, Li, Lin, and Cui]{ICTdecomposition}
Shao, S., Li, Y., Lin, Z., and Cui, Q.
\newblock High-rank irreducible cartesian tensor decomposition and bases of equivariant spaces.
\newblock \emph{Journal of Machine Learning Research}, 26\penalty0 (175):\penalty0 1--53, 2025.
\newblock URL \url{http://jmlr.org/papers/v26/25-0134.html}.

\bibitem[Su et~al.(2023)Su, Lu, Pan, Murtadha, Wen, and Liu]{RoPE}
Su, J., Lu, Y., Pan, S., Murtadha, A., Wen, B., and Liu, Y.
\newblock Roformer: Enhanced transformer with rotary position embedding, 2023.
\newblock URL \url{https://arxiv.org/abs/2104.09864}.

\bibitem[Team et~al.(2026)Team, Chen, Zhang, Su, Xu, Pan, Wang, Wang, Chen, Yin, et~al.]{AttnRes}
Team, K., Chen, G., Zhang, Y., Su, J., Xu, W., Pan, S., Wang, Y., Wang, Y., Chen, G., Yin, B., et~al.
\newblock Attention residuals.
\newblock \emph{arXiv preprint arXiv:2603.15031}, 2026.

\bibitem[Thomas et~al.(2018)Thomas, Smidt, Kearnes, Yang, Li, Kohlhoff, and Riley]{e3nn1}
Thomas, N., Smidt, T., Kearnes, S., Yang, L., Li, L., Kohlhoff, K., and Riley, P.
\newblock Tensor field networks: Rotation-and translation-equivariant neural networks for 3d point clouds.
\newblock \emph{arXiv preprint arXiv:1802.08219}, 2018.

\bibitem[Unke \& Maennel(2024)Unke and Maennel]{e3x}
Unke, O.~T. and Maennel, H.
\newblock \texttt{E3x}: $\mathrm{E}(3)$-equivariant deep learning made easy.
\newblock \emph{arXiv preprint arXiv:2401.07595}, 2024.

\bibitem[Vyas et~al.(2025)Vyas, Morwani, Zhao, Kwun, Shapira, Brandfonbrener, Janson, and Kakade]{SOAP}
Vyas, N., Morwani, D., Zhao, R., Kwun, M., Shapira, I., Brandfonbrener, D., Janson, L., and Kakade, S.
\newblock Soap: Improving and stabilizing shampoo using adam, 2025.
\newblock URL \url{https://arxiv.org/abs/2409.11321}.

\bibitem[Warford et~al.(2026)Warford, Thiemann, and Csányi]{without_U}
Warford, T., Thiemann, F.~L., and Csányi, G.
\newblock Better without u: Impact of selective hubbard u correction on foundational mlips, 2026.
\newblock URL \url{https://arxiv.org/abs/2601.21056}.

\bibitem[Weiler et~al.(2018)Weiler, Geiger, Welling, Boomsma, and Cohen]{e3nn2}
Weiler, M., Geiger, M., Welling, M., Boomsma, W., and Cohen, T.~S.
\newblock 3d steerable cnns: Learning rotationally equivariant features in volumetric data.
\newblock In Bengio, S., Wallach, H., Larochelle, H., Grauman, K., Cesa-Bianchi, N., and Garnett, R. (eds.), \emph{Advances in Neural Information Processing Systems}, volume~31. Curran Associates, Inc., 2018.

\bibitem[Wood et~al.(2026)Wood, Dzamba, Fu, Gao, Shuaibi, Barroso-Luque, Abdelmaqsoud, Gharakhanyan, Kitchin, Levine, Michel, Sriram, Cohen, Das, Rizvi, Sahoo, Ulissi, and Zitnick]{UMA}
Wood, B.~M., Dzamba, M., Fu, X., Gao, M., Shuaibi, M., Barroso-Luque, L., Abdelmaqsoud, K., Gharakhanyan, V., Kitchin, J.~R., Levine, D.~S., Michel, K., Sriram, A., Cohen, T., Das, A., Rizvi, A., Sahoo, S.~J., Ulissi, Z.~W., and Zitnick, C.~L.
\newblock Uma: A family of universal models for atoms, 2026.
\newblock URL \url{https://arxiv.org/abs/2506.23971}.

\bibitem[Xie et~al.(2026)Xie, Daigavane, Kotak, and Smidt]{vsh}
Xie, Y., Daigavane, A., Kotak, M., and Smidt, T.
\newblock Asymptotically fast clebsch-gordan tensor products with vector spherical harmonics, 2026.
\newblock URL \url{https://arxiv.org/abs/2602.21466}.

\bibitem[Xu et~al.(2026{\natexlab{a}})Xu, Wu, Xie, and Hu]{c3j}
Xu, Z., Wu, C., Xie, W., and Hu, P.
\newblock A cartesian-3j framework for machine learning interatomic potentials, 2026{\natexlab{a}}.
\newblock URL \url{https://arxiv.org/abs/2512.16882}.

\bibitem[Xu et~al.(2026{\natexlab{b}})Xu, Xie, and Hu]{TACE}
Xu, Z., Xie, W., and Hu, P.
\newblock Spectral/spatial tensor atomic cluster expansion with universal embeddings in cartesian space.
\newblock \emph{arXiv preprint arXiv:2509.14961}, 2026{\natexlab{b}}.

\bibitem[Yu et~al.(2026)Yu, Lin, Zhang, Qian, and Ji]{QHNetV2}
Yu, H., Lin, Y., Zhang, X., Qian, X., and Ji, S.
\newblock Efficient prediction of so(3)-equivariant hamiltonian matrices via so(2) local frames.
\newblock \emph{arXiv preprint arXiv:2506.09398}, 2026.

\bibitem[Zeng et~al.(2025)Zeng, Zhang, Peng, Zhang, He, Wang, Liu, Bi, Li, Cai, et~al.]{DeePMD-kit}
Zeng, J., Zhang, D., Peng, A., Zhang, X., He, S., Wang, Y., Liu, X., Bi, H., Li, Y., Cai, C., et~al.
\newblock Deepmd-kit v3: a multiple-backend framework for machine learning potentials.
\newblock \emph{Journal of chemical theory and computation}, 21\penalty0 (9):\penalty0 4375--4385, 2025.

\end{thebibliography}

\bibliographystyle{icml2026}


\newpage
\appendix
\onecolumn 

\newpage

\section{Training Details}
  All TECE models were trained in torch.float32 precision using 32 NVIDIA H20 GPUs. During pretraining, we employed DeNS~\cite{DeNS}and stochastic depth~\cite{stochastic_depth} as additional regularization techniques. DeNS was mainly used with the direct model to accelerate convergence, rather than to improve the final accuracy. With appropriately chosen hyperparameters, the direct and conservative models achieved comparable performance upon convergence. The DeNS hyperparameters were kept consistent with those used in EquiformerV3~\cite{eqv3}. For optimization, we used the Muon optimizer implemented in DeepMD-kit~\cite{Muon,DeePMD-kit,DPA4} to accelerate convergence. Muon is not essential. In fact, AdamW may be a better choice because it generally provides better extrapolation performance. However, considering the computational cost, we used Muon to accelerate convergence. Training the TECE-OAM-RRA-1.0 model consumed approximately 7680 NVIDIA H20 GPU-hours in total. This figure should not be interpreted as representative of the intrinsic computational efficiency of the TACE/TECE architecture, which achieves SOTA accuracy and computational performance. The computing platform used in our experiments exhibited suboptimal multi-GPU scaling and relied on outdated CUDA and PyTorch versions, preventing us from using compilation-based acceleration techniques. Consequently, the observed training cost was approximately three times higher than necessary. To provide a more representative evaluation, we benchmarked the computational performance of various TACE architectures on a newer platform. Further details on computational efficiency and inference speed are provided in the main text.

\begin{table}[h!]
  \caption{Hyperparameters used to train the TECE-OAM-RRA model series.}
  \centering
  \begin{tabular}{lccc}
  \toprule[1.2pt]
  Stage & Direct Pre-training & Conservative Pre-training & Fine-tuning  \\
  \midrule[1.2pt]
  Training Set & OMat24 & OMat24 & sAlex+8*MPtrj \\ 
  Optimizer & Muon & Muon & Muon \\ 
  Learning rate scheduling & WarmupStableDecay & WarmupStableDecay & WarmupStableDecay \\
  Number of Epoch & 1 & 2 & 2 \\
  Warmup Ratio & $10\%$ & $5\%$ & $5\%$ \\
  Stable Ratio & $90\%$ & $45\%$ & $45\%$ \\
  Decay Ratio & $0$ & $50\%$ & $50\%$ \\
  Stochastic Weight Averaging & False & False & True \\
  Maximum learning rate & $5 \times 10 ^{-3}$ & $3 \times 10 ^{-3}$ & $1 \times 10 ^{-4}$\\
  Minimum learning rate & - & $3 \times 10 ^{-6}$ & $1 \times 10 ^{-6}$\\
  Batch size & $512$ & $512$ & $256$ \\
  Weight decay & $1 \times 10 ^{-3}$ & $1 \times 10 ^{-3}$  & $1 \times 10 ^{-3}$\\ 
  Loss coefficient & 1/1/1 & 1/1/1  & 1/1/1\\
  Loss Type & MAE/L2MAE/MAE & MAE/L2MAE/MAE  & MAE/L2MAE/MAE\\
  Gradient clipping norm threshold & $100$ & $100$ & $100$\\
  Model EMA decay & $0.999$ & $0.999$ & $0.999$\\
  DeNS & True & False & False\\
  Stochastic Depth & 0.05 & 0 & 0\\
  \midrule[0.6pt]
  Cutoff radius ($\AA$) & $6$ & $6$ & $6$\\
  Number of layers & $8$ & $8$ & $8$\\
  Interaction channel & $64$ & $64$ & $64$\\
  Product channel & $256$ & $256$& $256$\\
  $L_{\mathrm{max}}$ & $4$ & $4$ & $4$\\
  $\ell_{\mathrm{max}}$ & $4$ & $4$ & $4$\\
  $m_{\mathrm{max}}$ & $4$ & $4$ & $4$\\
  $\ell_1\ell_2$ & $\leq$ & $\leq$ & $\leq$\\
  $m_1m_2$ & $\geq$ & $\geq$ & $\geq$\\
  Correlation order & $2$ & $2$ & $2$\\
  \bottomrule[1.2pt]
  \end{tabular}
  \vspace{1mm}
\end{table}

\newpage
\section{Matbench Discovery}
\begin{table}[h]
\caption{Unique Prototypes entries in Matbench Discovery at the time of uploading TECE-OAM-RRA-1.0.}
\label{tab:full_matbench}
\centering
\begin{small}
\scalebox{0.9}
{
\begin{tabular}{lccccccccccc}
\toprule[1.2pt]
Model & CPS & Acc & F1 & DAF & Prec & MAE & $R^2$ & $\kappa$SRME & RMSD & Params & Date Added \\
\midrule[1.2pt]
  TECE-OAM-RRA-1.0 & 0.908 &0.978 &	0.929	&6.073	&0.928 & 0.018 & 0.871 &	0.093 &	0.058	& 222M & 2026-07-05 \\
  EquFlashV2 & 0.907 & 0.978 & 0.929 & 6.069 & 0.928 & 0.018 & 0.873 & 0.094 & 0.058 & 44.9M & 2026-06-11 \\
  EquiformerV3+DeNS-OAM & 0.902 & 0.978 & 0.931 & 6.074 & 0.928 & 0.018 & 0.868 & 0.118 & 0.060 & 30.3M & 2026-04-07 \\
  PET-OAM-XL & 0.898 & 0.977 & 0.924 & 6.075 & 0.929 & 0.019 & 0.864 & 0.119 & 0.060 & 730M & 2026-01-10 \\
  GRACE-3L-OAM-L & 0.900	&0.977	&0.925&	6.041	&0.923	&0.018	&0.875	&0.121	& 0.058 & 42.1M &	2026-07-02 \\
  TACE-OAM-L & 0.889 & 0.972 & 0.910 & 5.898 & 0.902 & 0.020 & 0.868 & 0.126 & 0.061 & 82.9M & 2026-04-09 \\
  eSEN-30M-OAM & 0.888 & 0.977 & 0.925 & 6.069 & 0.928 & 0.018 & 0.866 & 0.170 & 0.061 & 30.2M & 2025-03-17 \\
  EquFlash & 0.888 & 0.975 & 0.919 & 5.983 & 0.915 & 0.019 & 0.871 & 0.158 & 0.060 & 28.7M & 2025-06-23 \\
  Nequip-OAM-XL & 0.886 & 0.971 & 0.906 & 5.869 & 0.897 & 0.020 & 0.872 & 0.125 & 0.063 & 32.1M & 2025-11-30 \\
  MatRIS-10M-OAM & 0.877 & 0.976 & 0.921 & 6.039 & 0.923 & 0.019 & 0.871 & 0.218 & 0.060 & 10.4M & 2025-10-29 \\
  SevenNet-Omni-i12 & 0.873 & 0.971 & 0.906 & 5.954 & 0.910 & 0.021 & 0.868 & 0.192 & 0.062 & 54.9M & 2026-01-12 \\
  Nequip-OAM-L & 0.870 & 0.967 & 0.893 & 5.823 & 0.890 & 0.022 & 0.865 & 0.166 & 0.065 & 9.6M & 2025-09-08 \\
  GRACE-2L-OAM-L & 0.865 & 0.964 & 0.883 & 5.840 & 0.893 & 0.022 & 0.862 & 0.169 & 0.064 & 26.4M & 2025-09-09 \\
  ORB v3 & 0.860 & 0.971 & 0.905 & 5.912 & 0.904 & 0.024 & 0.821 & 0.210 & 0.075 & 25.5M & 2025-04-05 \\
  DPA-4.0.1-Pro-MPtrj & 0.840 & 0.956 & 0.857 & 5.609 & 0.857 & 0.029 & 0.836 & 0.211 & 0.069 & 22.8M & 2026-06-11 \\
  Allegro-OAM-L & 0.840 & 0.966 & 0.895 & 5.674 & 0.867 & 0.022 & 0.868 & 0.319 & 0.065 & 9.7M & 2025-09-08 \\
  GRACE-2L-OAM & 0.837 & 0.963 & 0.880 & 5.774 & 0.883 & 0.023 & 0.862 & 0.294 & 0.067 & 12.6M & 2025-02-06 \\
  EquiformerV3+DeNS-MP & 0.830 & 0.956 & 0.863 & 5.479 & 0.838 & 0.029 & 0.840 & 0.275 & 0.070 & 30.3M & 2026-04-07 \\
  DPA-3.1-3M-FT & 0.802 & 0.963 & 0.884 & 5.667 & 0.866 & 0.023 & 0.869 & 0.470 & 0.069 & 3.27M & 2025-06-05 \\
  eSEN-30M-MP & 0.797 & 0.946 & 0.831 & 5.260 & 0.804 & 0.033 & 0.822 & 0.340 & 0.075 & 30.1M & 2025-03-17 \\
  MACE-MPA-0 & 0.795 & 0.954 & 0.852 & 5.582 & 0.853 & 0.028 & 0.842 & 0.412 & 0.073 & 9.06M & 2024-12-09 \\
  MatRIS-10M-MP & 0.778 & 0.951 & 0.847 & 5.422 & 0.829 & 0.031 & 0.824 & 0.489 & 0.072 & 10.4M & 2025-10-29 \\
  AlphaNet-v1-OAM & 0.769 & 0.968 & 0.901 & 5.747 & 0.879 & 0.024 & 0.831 & 0.644 & 0.079 & 4.65M & 2025-05-12 \\
  MatterSim v1 5M & 0.767 & 0.959 & 0.862 & 5.852 & 0.895 & 0.024 & 0.863 & 0.575 & 0.073 & 4.55M & 2024-12-16 \\
  GRACE-1L-OAM & 0.761 & 0.944 & 0.824 & 5.255 & 0.803 & 0.031 & 0.842 & 0.517 & 0.072 & 3.45M & 2025-02-06 \\
  Eqnorm MPtrj & 0.756 & 0.929 & 0.786 & 4.844 & 0.741 & 0.040 & 0.799 & 0.408 & 0.084 & 1.31M & 2025-05-26 \\
  Nequix MP PFT & 0.755 & 0.914 & 0.748 & 4.479 & 0.685 & 0.044 & 0.784 & 0.307 & 0.087 & 708k & 2026-01-08 \\
  Nequip-MP-L & 0.733 & 0.921 & 0.761 & 4.704 & 0.719 & 0.043 & 0.791 & 0.452 & 0.086 & 9.6M & 2025-09-08 \\
  Nequix MP & 0.729 & 0.914 & 0.751 & 4.455 & 0.681 & 0.044 & 0.782 & 0.446 & 0.085 & 708k & 2025-08-17 \\
  Allegro-MP-L & 0.720 & 0.915 & 0.751 & 4.516 & 0.690 & 0.044 & 0.778 & 0.504 & 0.082 & 18.7M & 2025-09-08 \\
  SevenNet-l3i5 & 0.714 & 0.920 & 0.760 & 4.629 & 0.708 & 0.044 & 0.776 & 0.550 & 0.085 & 1.17M & 2024-12-10 \\
  HIENet & 0.707 & 0.929 & 0.777 & 4.932 & 0.754 & 0.041 & 0.793 & 0.642 & 0.080 & 7.51M & 2025-07-01 \\
  GRACE-2L-MPtrj & 0.681 & 0.895 & 0.691 & 4.163 & 0.636 & 0.052 & 0.741 & 0.526 & 0.090 & 15.3M & 2024-11-21 \\
  MACE-MP-0 & 0.637 & 0.878 & 0.669 & 3.777 & 0.577 & 0.057 & 0.697 & 0.682 & 0.092 & 4.69M & 2023-07-14 \\
  eqV2 M & 0.558 & 0.975 & 0.917 & 6.047 & 0.924 & 0.020 & 0.848 & 1.771 & 0.069 & 86.6M & 2024-10-18 \\
  ORB v2 & 0.528 & 0.965 & 0.880 & 6.041 & 0.924 & 0.028 & 0.824 & 1.734 & 0.097 & 25.2M & 2024-10-11 \\
  eqV2 S DeNS & 0.522 & 0.939 & 0.815 & 5.042 & 0.771 & 0.036 & 0.788 & 1.676 & 0.076 & 31.2M & 2024-10-18 \\
  ORB v2 MPtrj & 0.470 & 0.922 & 0.765 & 4.702 & 0.719 & 0.045 & 0.756 & 1.726 & 0.101 & 25.2M & 2024-10-14 \\
  M3GNet & 0.428 & 0.812 & 0.569 & 2.882 & 0.441 & 0.075 & 0.585 & 1.409 & 0.112 & 228k & 2022-09-20 \\
  CHGNet & 0.400 & 0.851 & 0.613 & 3.361 & 0.514 & 0.063 & 0.689 & 1.717 & 0.095 & 413k & 2023-03-03 \\
  GNoME & n/a & 0.948 & 0.829 & 5.523 & 0.844 & 0.035 & 0.785 & n/a & n/a & 16.2M & 2024-02-03 \\
\bottomrule[1.2pt]
\end{tabular}
}
\end{small}
\vspace{1mm}
\end{table}


\end{document}